\documentclass[10pt,twocolumn,letterpaper]{article}

\usepackage{cvpr}
\usepackage{times}
\usepackage{epsfig}
\usepackage{graphicx}
\usepackage{amsmath}
\usepackage{amssymb}
\usepackage{algorithm2e}
\usepackage{multirow}
\usepackage{tikz, subfigure, multirow}
\usepackage{color}
\usepackage{xcolor}
\usepackage{mathtools, nccmath}
\usepackage{hhline}
\usepackage{makecell}
\usepackage{bm}

\usepackage{comment}
\usepackage{tabularx}
\usepackage{float}
\usepackage{stfloats}
\usepackage{diagbox}
\usepackage{indentfirst}
\usepackage{enumitem}
\usepackage{algorithmicx}
\usepackage{pifont}
\usepackage{subfiles}

\usepackage[pagebackref=true,breaklinks=true,letterpaper=true,colorlinks,bookmarks=false]{hyperref}

\cvprfinalcopy 

\def\cvprPaperID{2473} 

\ifcvprfinal\fi
\begin{document}

\title{DMCP: Differentiable Markov Channel Pruning for Neural Networks}

\author{Shaopeng Guo \quad Yujie Wang \quad Quanquan Li \quad Junjie Yan \\
SenseTime Research\\
{\tt\small \{guoshaopeng, wangyujie, liquanquan, yanjunjie\}@sensetime.com}}

\maketitle

\begin{abstract}
    Recent works imply that the channel pruning can be regarded as searching optimal sub-structure from unpruned networks.
    However, existing works based on this observation require training and evaluating a large number of structures, which limits their application.
    In this paper, we propose a novel differentiable method for channel pruning, named Differentiable Markov Channel Pruning (DMCP), to efficiently search the optimal sub-structure. 
    Our method is differentiable and can be directly optimized by gradient descent with respect to standard task loss and budget regularization (e.g. FLOPs constraint). 
    In DMCP, we model the channel pruning as a Markov process, in which each state represents for retaining the corresponding channel during pruning, and transitions between states denote the pruning process.
    In the end, our method is able to implicitly select the proper number of channels in each layer by the Markov process with optimized transitions. To validate the effectiveness of our method, we perform extensive experiments on Imagenet with ResNet and MobilenetV2.
    Results show our method can achieve consistent improvement than state-of-the-art pruning methods in various FLOPs settings. The code is available at~\url{https://github.com/zx55/dmcp}
\end{abstract}

\section{Introduction}

\label{sec:introduction}

Channel pruning~\cite{li2016pruning,Cun90optimalbrain,Hassibi:1993:OBS:2987189.2987223} has been widely used for model acceleration and compression.
The core idea behind is that large CNN models are regarded as over-parameterized. 
By removing the large model's unnecessary or less important weights, we can obtain a more efficient and compact model with a marginal performance drop.
Conventional channel pruning methods mainly rely on the human-designed paradigm.
A typical pipeline of conventional pruning method can be summarized as three stages: pre-train a large model, prune ``unimportant'' weights of the large model according to the pre-defined criterion, fine-tune the pruned model~\cite{he2018progressive,liu2017learning,molchanov2016pruning, DBLP:journals/corr/LiKDSG16}. 

\begin{figure}[t!]
    \centering
    \includegraphics[width=0.4\textwidth]{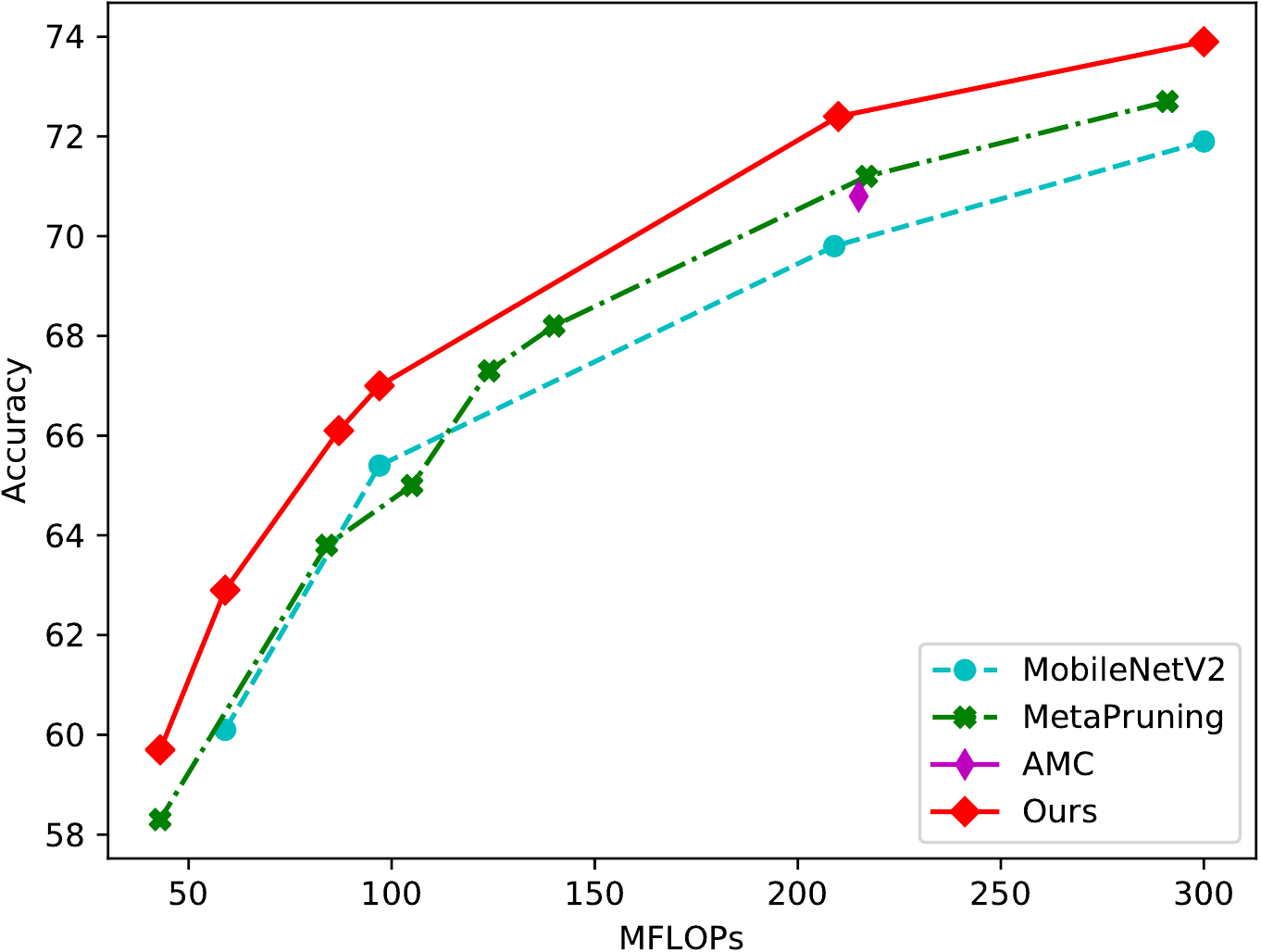}
    \caption{MFLOPs vs. Accuracy on the ImageNet classification dataset. The original model is MobileNetV2~\cite{sandler2018mobilenetv2}. Our method outperforms existing pruning methods MetaPruning~\cite{liu2019metapruning} and AMC~\cite{he2018amc} on mobile settings ($<$300MFLOPs) at all FLOPs. See Table~\ref{tb:models} for more results. Best viewed in color.
    \vspace{-1em}}
    \label{fig:figure1}
\end{figure}

Recent work~\cite{liu2018rethinking} showed a new perspective of channel pruning that 
the structure of the pruned model is the key of determining the performance of a pruned model, rather than the inherited ``important'' weights.
Based on this observation, some works try to design a pruning process to directly search optimal sub-structure from the unpruned structure.
AMC~\cite{he2018amc} adopted reinforcement learning (RL) to train a controller to output the pruning ratio of each layer in the unpruned structure, while MetaPruning~\cite{liu2019metapruning} used evolution algorithm to search structures. However, the optimization of these pruning process need to train and evaluate a large number of structures sampled from the unpruned network, thus the scalability of these methods is limited.
Although AMC don't fine-tune the pruned structures and MetaPruning trained a meta-network to predict network's weights to avoid training the searched structures, the limitation of scalability still remains.

A similar problem in neural architecture search (NAS) has been tackled by differentiable method DARTS~\cite{liu2018darts}.
However, the differentiable method proposed by DARTS cannot be directly applied to channel pruning.
First, the definition of search space is different.
The search space of DARTS is a category of pre-defined operations (convolution, max-pooing, etc), while in the channel pruning, the search space is the number of channels in each layer.
Second, the operations in DARTS are independent with each other.
But in the channel pruning,
if a layer has $k+1$ channels, it must have at least $k$ channels first, which has a logical implication relationship.

In this paper, we propose a novel differentiable channel pruning method named Differentiable Markov Channel Pruning (DMCP) to perform efficient optimal sub-structure searching.
Our method makes the channel pruning differentiable by modeling it as a Markov process.
In the Markov process for each layer, the state $S_k$ represents the $k^{th}$ channel is retained, and the transition from $S_k$ to $S_{k+1}$ represents the probability of retaining the $(k+1)^{th}$ channel given that the $k^{th}$ channel is retained.
Note that the start state is always $S_1$ in our method.
Then the marginal probability for state $S_k$, i.e. the probability of retaining $k^{th}$ channel, can be computed by the product of transition probabilities and can also be viewed as a scaling coefficient.
Each scaling coefficient is multiplied to its corresponding channel's feature map during the network forwarding.
So the transition probabilities parameterized by learnable parameters can be optimized in an end-to-end manner by gradient descent with respect to task loss together with budget regularization (e.g. FLOPs constraint).
After the optimization, the model within desired budgets can be sampled by the Markov process with learned transition probabilities and will be trained from scratch to achieve high performance.
The details of our design will be presented in Section~\ref{sec:method}.

Finally, to demonstrate the effectiveness of our method, we conduct exhaustive classification experiments on ImageNet~\cite{ILSVRC15}.
At the same FLOPs, our method outperforms all the other pruning methods both on MobileNetV2 and ResNet, as shown in Figure~\ref{fig:figure1}. 
With our method, MobileNetV2 has 0.1\% accuracy drop with 30\% FLOPs reduction and the FLOPs of ResNet-50 is reduced by 44\% with only 0.4\% drop. 

\section{Related Work}
\label{sec:related_work}

In this section, we discuss related works from network architecture search (NAS) and channel pruning.

\noindent\textbf{Neural Architecture Search.}
\cite{zoph2016neural} first proposed to search for neural architectures with reinforcement learning to achieve competitive accuracy with the given inference cost.
But the searching cost is too expensive to be applied broadly.
Recent works try to reduce the searching cost by gradient-based methods.
DARTS~\cite{liu2018darts} used a set of learnable weights to parameterize the probabilities of each candidate operation, the output of a layer is the linear combination of probabilities and feature maps of corresponding operation. After training, the operation with the highest probability is chosen to be the final architecture. However, DARTS is performed on a small proxy task (e.g. CIFAR10) and transfer the searched architecture to large scale target tasks (e.g. ImageNet).
ProxylessNAS~\cite{cai2018proxylessnas} avoided using proxy tasks by only sampling two paths to search for architecture on large scale target tasks.
Different from searching architecture with different types of operations in the NAS methods mentioned above, our method focuses on searching structures with a different number of channels.

\noindent\textbf{Channel Pruning.}
Previous works on channel pruning can be roughly classified into two categories, i.e. hard pruning and soft pruning.
Hard pruning removes channels during iterative pruning and fine-tuning process, while soft pruning only makes the pruned channels to be or approach to zero.
Hard pruning methods mainly depend on different pruning criteria, 
for example, weight norm~\cite{li2016pruning}, the average percentage of zeros in the output ~\cite{hu2016network} or the influence of each channel to the final loss~\cite{molchanov2016pruning}. 
Soft pruning methods mainly make the pruned channels to be or approach to zero so that those channels' influence is decreased.
\cite{he2018progressive} first zero some filters by intra-layer criterion and a calculated layer-wise ratio.
Then it increases the ratio of pruned filters gradually until reaching the given computation budget. 
\cite{liu2017learning} add L1 regularization on Batch Normalization's coefficients when training, and after training the channels with small coefficients will be pruned.
\cite{DBLP:journals/corr/abs-1905-04748} search for the least important filters in a binary search manner.
\cite{DBLP:journals/corr/abs-1903-09291} use  generative adversarial learning to learn a sparse soft mask to scaled the output of pruned filters toward zero.

Our method can be seen as soft pruning.
The major difference among DMCP and the above methods is the elimination of duplicated solutions by our Markov modeling.
For example, given a layer with $C$ channels, the solution space of our method is $O(C)$, but for methods mentioned above, the solution space is $O(2^C)$ for different combinations even with the same number of channels.

Based on recent work~\cite{liu2018rethinking}, some work designed a search process to directly search the optimal sub-structures from the unpruned net.
AMC~\cite{he2018amc} used reinforcement learning to determinate the ratio of channels each layer should retain. 
MetaPruning~\cite{liu2019metapruning} used an evolution algorithm to search network structures and a meta network is trained to predict weights for network structures during searching.
These methods need to train or evaluate a large number structures, which makes them inefficient, while our method can be optimized by gradient descent and avoid the problem.

\section{Method}
\label{sec:method}
\begin{figure*}[t!]
    \centering
    \includegraphics[width=0.97\textwidth]{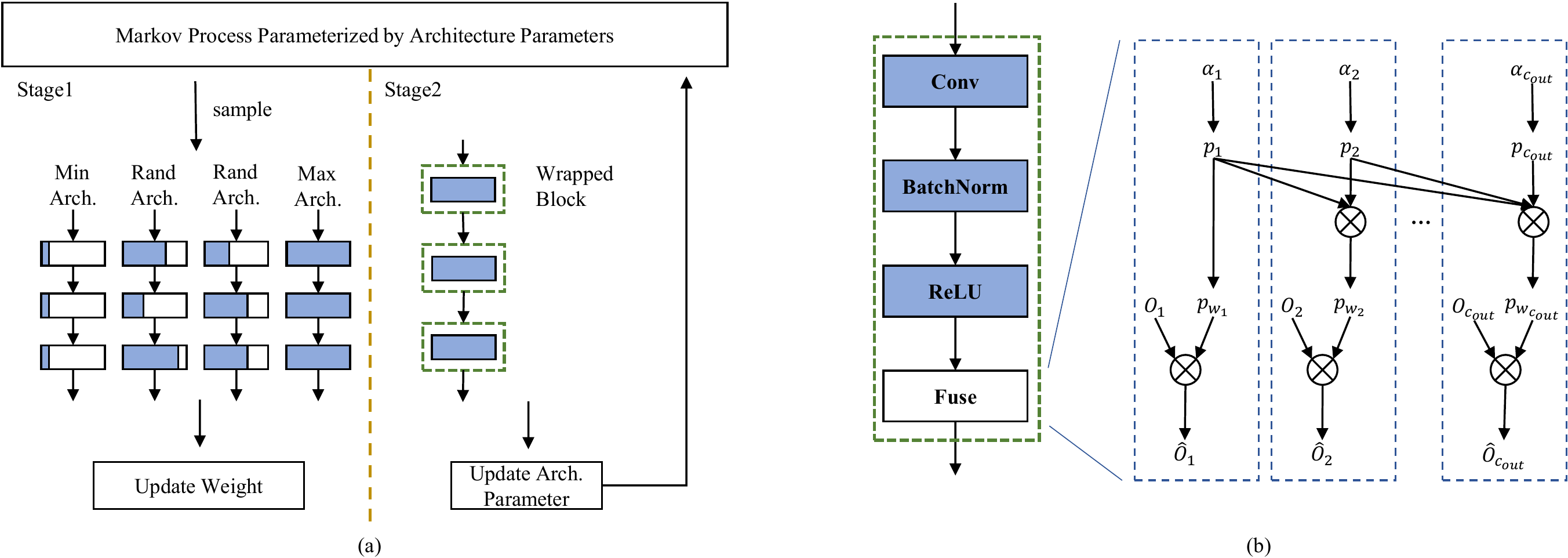}
    \caption{The training pipeline of DMCP. Figure (a) demonstrates the two stages of DMCP. DMCP first run stage1 for several iterations to update weights of the unpruned network to warmup, then run stage1 and stage2 iteratively to update weights and architecture parameters. In figure (a), each rectangle represents a convolution block, e.g. Conv-BN-ReLU. 
    Four sub-structures, represented by the blue parts of the rectangle, are sampled from the unpruned net: (1) the whole unpruned net (Max. Arch.), (2) structure with the minimum number of channels (Min. Arch.), (3) two structures randomly sampled by Markov process (Rand. Arch.). 
    Each of these structures is forwarded independently, and the gradients in four sub-structure are accumulated to update the weights.
    Figure (b) is a detail illustration of the wrapped block in figure (a). The ``Fuse'' layer shows the incorporate details of architecture parameters $\alpha$ and outputs of unpruned networks $O$. Notations in Figure (b) are explained in Section~\ref{sec:method}. Best viewed in color.
    \vspace{-1em}}
    \label{fig:pipeline}
\end{figure*}

In this section, we will give a detailed explanation of the proposed Differentiable Markov Channel Pruning (DMCP) method.
As illustrated in Section~\ref{sec:pruning_process},
the channel pruning is first formulated as a Markov process parameterized by architecture parameters and can be optimized in an end-to-end manner.
Then in Section~\ref{sec:hpn_train},
the training procedure of DMCP can be divided into two stages: in stage 1, the unpruned network is updated by our proposed \textbf{variant sandwich rule}, while in stage 2, the architecture parameters are wrapped into the unpruned network and get updated, as shown in Figure~\ref{fig:pipeline} (a).
After the optimization, we propose two ways to sample the pruned network in Section~\ref{sec:produce_model}.

\subsection{Definition of Pruning Process}
\label{sec:pruning_process}
Let $M(L^{(1)}, L^{(2)}, ..., L^{(N)})$ denote the $N$-layer unpruned network, where $L^{(i)}$ is the $i^{th}$ layer. In layer $L^{(i)}$ with $C_{out}^{(i)}$ convolutional filters (i.e. channels), given input $x$, the output $O^{(i)}$ can by computed by:
\begin{align}
\label{eq:conv}
O_{k}^{(i)} = w_k^{(i)} \odot x, k = 1, 2, ..., C_{out}^{(i)}
\end{align}
where $O_k^{(i)}$ is the $k^{th}$ channel of $O^{(i)}$, $w_k^{(i)}$ is the $k^{th}$ filter in $L^{(i)}$, and $\odot$ denote the convolution operation. 
If not explicitly stated, the superscript which represents the layer index will be omitted below for simplicity.

As illustrated in Section~\ref{sec:introduction}, we perform the channel pruning in a reversed way, which can be represented by a directed ascyclic graph, as shown in Figure~\ref{fig:markov_chain}, where the state $S_k(1 \leq k \leq C_{out})$ represent $k^{th}$ channel is retained during the pruning process, and the transition $p_k$ from $S_k$ to $S_{k+1}$ means $(k+1)^{th}$ channel is retained if $k^{th}$ channel is retained. The pruning process can be ended by transferring to the terminal state $T$ from any other state. 
This process has the property that if $k$ out of $C_{out}$ channels are retained in layer $L$, they must be first $k$ channels. In other words, given $k^{th}$ channel is retained, then first $(k-1)$ channels must be retained, and we can further conclude that retaining $(k+1)^{th}$ channel is conditional independent of first $(k-1)$ channels give $k^{th}$ channel is retained, which follows the Markov property.

\subsubsection{Channel Pruning via Markov Process}
\label{sec:relaxation}
We model transition in aforementioned ascyclic graph as a stochastic process and parameterized the transition probabilities by a set of learnable parameters.
We name the learnable parameters as \textbf{architecture parameters} for distinguishing them from network weights.
Let $p(w_1, w_2, ..., w_{k-1})$ be the probability that first $k - 1$ channels are retained. The probability of retaining first $k$ channels can be represented as:
\begin{align}
\label{eq:first_k}
p(w_1, ..., w_k) = p(w_k|w_1, ..., w_{k-1})p(w_1, ..., w_{k-1})
\end{align}
where $p(w_k|w_1, ..., w_{k-1})$ is the probability of retaining $k^{th}$ channel given first $(k-1)$ channels are retained.
Since retaining $w_k$ is conditionally independent of $\{w_1, w_2, ... w_{k-2}\}$ given $w_{k-1}$ is retained, hence we can rewrite Equation~\ref{eq:first_k} as:
\begin{align}
\label{eq:condition_first_k}
p_k = p(w_k |w_1, w_2, ..., w_{k-1}) = p(w_k |w_{k-1})
\end{align}
\begin{align}
\label{eq:condition_not_first_k}
p(w_k |\neg w_{k-1}) = 0
\end{align}
in which $\neg w_{k-1}$ means $(k-1)^{th}$ channel is discarded. 

\begin{figure}[h]
    \centering
    \includegraphics[width=0.4\textwidth]{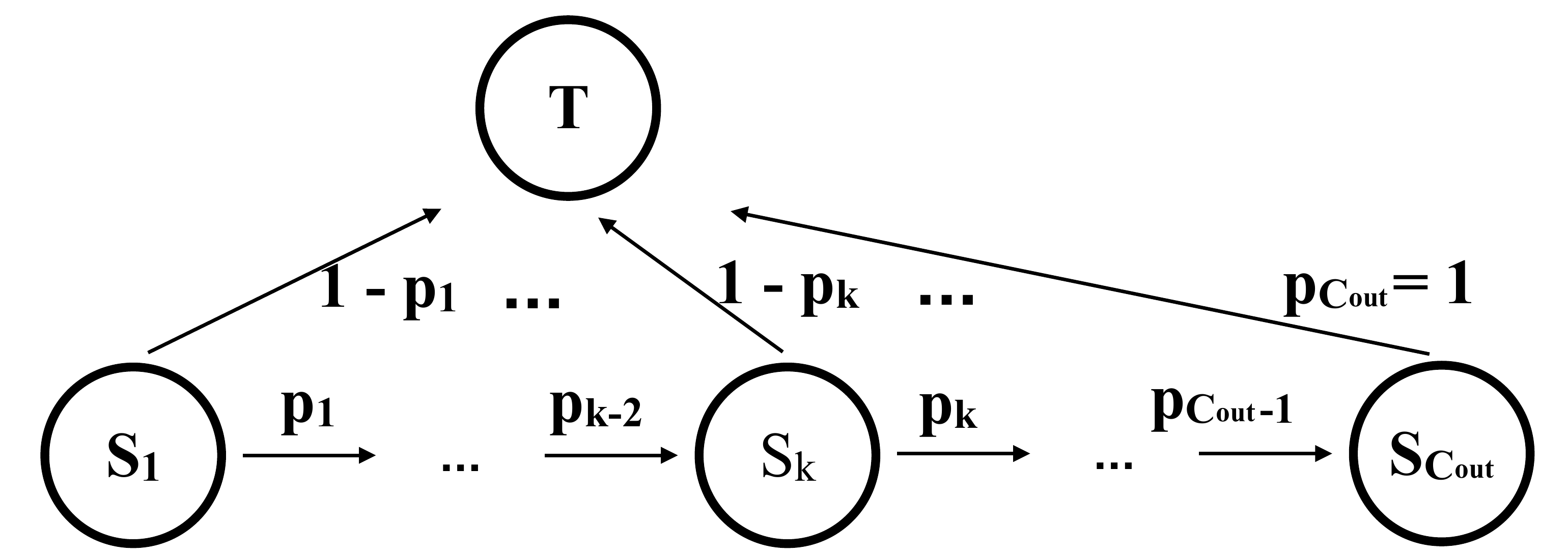}
    \caption{The Modeling of channel pruning as a Markov process. State $S_k(k=1, 2, ....)$ means $k^{th}$ channel is retained, and transition $p_k$ is the probability of retaining $k^{th}$ channel given $(k-1)^{th}$ channel is retained, while $1-p_k$ is the probability of terminating the process. State $T$ means the terminal state and $C_{out}$ is the maximum number of channels in each layer.}
    \label{fig:markov_chain}
\end{figure}

Therefore, in Figure~\ref{fig:markov_chain}, the transitions can be represented by a set of transition probabilities $P=\{p_1, p_2, .. p_{C_{out}}\}$ that defined by  Equation~\ref{eq:condition_first_k}.

We use a set of architecture parameters $A = \{\alpha_1, \alpha_2, ..., \alpha_{c_{out}}\}$ to parameterize $P$, therefore $p_k$ can be computed as follows:
\begin{align}
p_k = 
\begin{cases} 
\label{eq:prob}
      1 & k = 1 \\
      sigmoid(\alpha_k) = \frac{1}{1 + e^{-\alpha_k}} & k = 2, ..., C_{out}, \alpha_k \in A
\end{cases}
\end{align}

Note that we leave at least one channel for each layer, so $p_1 = p(w_1) = 1$.

The marginal probability of sampling channel $w_k$ denoted by $p(w_k)$ can be computed as:
{\small
\begin{equation}
\begin{aligned}
\label{eq:marginal_prob}
    p(w_k) &= p(w_k|w_{k-1})p(w_{k-1}) + p(w_k|\neg w_{k-1})p(\neg  w_{k-1}) \\
    &= p(w_k|w_{k-1})p(w_{k-1}) + 0\\
    &= p(w_1)\prod_{i=2}^{k} p(w_i|w_{i-1}) = \prod_{i=1}^{k} p_i 
\end{aligned}
\end{equation}
}
Then the architecture parameters are wrapped into the unpruned network by following equation: 
\begin{align}
\label{eq:upd_a_fwd}
\hat{O_k} = O_k \times p(w_k)
\end{align}
where $\hat{O_{k}}$ is the actual output of $k^{th}$ channel. Therefore, pruning $w_k$ can be represented by setting $p(w_k)$ to zero.

However, we cannot directly implement Equation~\ref{eq:upd_a_fwd} right after the convolutional layer, because the batch normalization layer can scale up the value of $i^{th}$ channel such that the latter layer will not be affected.
So the pruning process should be put after the batch normalization layer.
An example of how to combine architecture parameters with an unpruned network is given in Figure~\ref{fig:pipeline} (b).

By the above definition, the pruned model can be sampled by the Markov process, while the transitions can be optimized by gradient descent, which will be illustrated in Section~\ref{sec:hpn_train}.

\subsubsection{Solution to Shortcut Issue}
Note that both MobilenetV2 and ResNet have residual blocks with shortcut connections.
For the residual blocks with identity shortcuts, the number of channels in the last convolutional layer must be the same as the one in previous blocks due to the element-wise summation. 
Many previous works~\cite{li2016pruning, hu2016network} don't prune the last convolutional layer of the residual block.
In our method, we adopt a weight sharing strategy to solve this issue such that the layers, whose output channels must be equal, will share the same set of architecture parameters.

\subsubsection{Budget Regularization}
FLOPs and latency are commonly used in evaluating pruning methods.
To perform an easy-to-implement and fair comparison, we use accuracy at certain FLOPs as budget regularization.
However, budget regularization like FLOPs cannot be naturally optimized by gradient descent.
In this section, we introduce our solution to handle the non-differentiable budget regularization problem.

In layer $L$, the expected channel $E(channel)$ can be computed as:
\begin{align}
\label{eq:e_channel}
E(channel) = \sum_{i=1}^{C_{out}}p(w_i)
\end{align}
where $C_{out}$ is the number of output channels in $L$ and $p(w_i)$ is the marginal probability defined in Equation~\ref{eq:marginal_prob}. 

In layer $L$, given expected input channels $E(in)$ and output channels $E(out)$ computed as Equation~\ref{eq:e_channel}, the expected FLOPs $E(L_{FLOPs})$ can be computed by:
\begin{eqnarray}
\label{eq:e_fl_layer}
E(L_{FLOPs}) =& E(out) \times E(kernel\_op) \\
E(kernel\_op) =& \frac{E(in)}{groups} \times \#channel\_op \\
\#channel\_op =& (\frac{S_I + S_P - S_K}{stride} + 1) \times S_K \times S_K
\end{eqnarray}
where $groups=1$ for normal convolution and $groups=E(in)$ for depth-wise convolution. 
$S_I$ and $S_K$ indicate input width/height and kernel width/height respectively, while $S_P$ is padding size and $stride$ is convolution stride.

Then the expected flops of the model $E(N_{FLOPs})$ is:
\begin{align}
\label{eq:e_flops}
E(N_{FLOPs}) = \sum_{l=1}^{N}E^{(l)}(L_{FLOPs})
\end{align}
in which N is the number of convolutional layers.
With Equation~\ref{eq:e_flops}, we can optimize FLOPs by gradient descent.

\subsubsection{Loss Function}
Given target FLOPs $FLOPs_{target}$, we formulated the differentiable budget regularization loss $loss_{reg}$ as follows:
\begin{align}
\label{eq:flops_loss}
loss_{reg} = log(|E(N_{FLOPs}) - FLOPs_{target}|)
\end{align}
To make $E(N_{FLOPs})$ strictly lower than $FLOPs_{target}$ but not too sensitive around the target, we add single side margin to the loss function, i.e. when $\gamma \times FLOPs_{target} \leq E(N_{FLOPs}) \leq FLOPs_{target}$ is satisfied, the loss will be zero. $\gamma < 1$ is the tolerance ratio that can be adjusted by users.

When updating weights, the FLOPs loss has no effect on weights, so the loss function is:
\begin{align}
\label{eq:upd_w_loss}
Loss_{weight} = loss_{cls}
\end{align}
where $loss_{cls}$ is cross entropy loss for classification.
When updating the architecture parameters, the loss function is formulated as below:
\begin{align}
\label{eq:upd_a_loss}
Loss_{arch} = loss_{cls} + \lambda_{reg}loss_{reg}
\end{align}
where $\lambda_{reg}$ is hyper-parameters to balance two loss terms.

Note that we don't add weight decay to architecture parameters.
Because when the probability of keeping some channels approaching to zero or one, the norm of learnable parameters $\alpha$ will become very large, which will make them move forward to zero and hinge the optimization.

\subsection{Training Pipeline}
\label{sec:hpn_train}
As illustrated in Figure~\ref{fig:pipeline} (a), the training procedure of DMCP can be divided into two stages, i.e. weight updating of the unpruned network and architecture parameters updating.
The stage 1 and stage 2 are called iteratively during the training.

\noindent\textbf{Stage 1: Weight updating of the unpruned network.}
In the first stage, we only update weights in the unpruned network.
As defined in Equation~\ref{eq:marginal_prob}, the probability of retaining the $k^{th}$ channel can also be regarded as the probability of retaining the first $k$ channels.
Then our method can be seen as soft sampling all sub-structures in a single forwarding when updating architecture parameters.
 In general, all channels in a layer are equal and it is not intuitive to modeling the channel selection as Markov process. Inspired by previous work~\cite{yu2019universally}, which proposed a ``sandwich rule''  training method that the  $0.75\times$ parts of the trained MobileNetV2 $1.0\times$ can get similar performance to it trained from scratch.
, we introduce a \textbf{variant sandwich rule}, into the training scheme to make the channel groups in the unpruned model more ``important'' than the channel groups right after it.
So that channels in a layer are not equal.
The best choice of a layer with $k$ channels will be the first $k$ channels instead of other possible combinations.
Based on this channel importance ranking property in the unpruned model, when sampling a sub-network with $k(k<C)$ channels, selecting the first $k$ channels can better indicate the true performance of the sub-network (trained from scratch individually). 
Therefore, it is reasonable to introduce Markov modeling.

There are two differences between our variation and the original ``sandwich rule".
First, the randomly sampled switch (the ratio of retained channels) in each layer is not the same.
Because the pruned network may have different switches in different layers.
Second, the random sampling of switches obeys distribution from architecture parameters with the Markov process, instead of uniform distribution.
Because the possible number of architecture in our method is much more than \cite{yu2019universally}.
And to make all architectures reflect their true performance will need too much costs.
Thus we only focus on the frequently sampled architectures.

\noindent\textbf{Stage 2: Architecture parameter updating.}
In the second stage, we only update architecture parameters.
For each convolutional layer in the unpruned net, an architecture parameter is incorporated with its original output tensors by Equation~\ref{eq:upd_a_fwd}.
So that gradients could be backpropagated to the architecture parameters.
And the gradients will be backpropagated to $\alpha$ by following formulas:
\begin{gather}
\label{eq:upd_a_bwd}
\frac{\partial Loss}{\partial \alpha^{(i)}_j} = \sum_{k=1}^{C_{out}} \frac{\partial Loss}{\partial \hat{O^{(i)}_{k}}} \times \frac{\partial \hat{O^{(i)}_k}}{\alpha^{(i)}_j} \\
{\small
\frac{\partial \hat{O^{(i)}_k}}{\alpha^{(i)}_j}=
\begin{cases} 
      0 & ,k < j \\
     \frac{\partial p_k}{\partial \alpha_j} O^{(i)}_k \prod_{r \in \{r|r \neq j \, and \, r \leq k\}} p_r   & ,k \ge j
\end{cases}} \\
\frac{\partial p_k}{\partial \alpha_j} = (1 - p_k) p_k
\end{gather}
Such that all components of our method can be trained in an end-to-end manner.
To further reduce the search space, we divide the channels into groups ($\geq10$ groups) uniformly and each architecture parameter $\alpha$ is responsible for one group instead of only one channel.
Each layer has the same number of groups.

\noindent\textbf{Warmup process.}
Before iteratively called stage 1 and stage 2, DMCP first runs stage 1 for several epochs  to warm up, in which the sub-networks are sampled by Markov process with randomly initialized architecture parameters.
This process aims to avoid the network dropping into bad local minima when updating architecture parameters caused by weights' insufficient training. We also conduct ablation study in section~\ref{sec:ablation} to show the effectiveness of using warmup.

\subsection{Pruned Model Sampling}
\label{sec:produce_model}
After DMCP training done, we then produce models that satisfy the given cost constrain from it.
In this section, we will introduce two producing methods.
The first method, named \textbf{Direct Sampling (DS)}, is to sample in each layer independently by the Markov process with optimized transition probabilities.
We sample several structures and only keep the structures that lie in the target FLOPs budget.

The second method, named \textbf{Expected Sampling (ES)}, is to set the number of channels in each layer to be the expected channels computed by Equation~\ref{eq:e_channel}.
In our experiment, $loss_{reg}$ is always optimized to zero, so the FLOPs of the expected network is equal or less than the given FLOPs constraint.
Thus the expected network also satisfies the requirements.

In Section~\ref{sec:exp}, we perform plenty of experiments to compare these two methods.
The best performance of the pruned model sampled from Direct Sampling is a little bit higher than the one produced by Expected Sampling method, but it takes a much longer time to find such a model.
So in our experiments, we use the Expected Sampling method to produce the final pruned model.

\section{Experiments}
\label{sec:exp}
In this section, we perform a large number of experiments to validate and analyze our method.
We first describe the implementation details of DMCP in Section~\ref{sec:imp_detail}.
To study the effectiveness of each component in our method, we conduct ablation experiments in 
Section~\ref{sec:ablation}.
Finally in Section~\ref{sec:compare}, we compare our results with state-of-the-art channel pruning methods.
More visualizations and experiments will be shown on Supplemental Materials.

\subsection{Implementation Details}
\label{sec:imp_detail}
We demonstrate the effectiveness of our proposed differentiable pruning method on ImageNet classification~\cite{russakovsky2015imagenet}, which contains 1000 classes. 
We perform experiments on both light (MobileNetV2~\cite{DBLP:journals/corr/abs-1801-04381}) and heavy (ResNet~\cite{DBLP:journals/corr/HeZRS15}) models.
For MobilenetV2, we use MobilenetV2 1.5x as the unpruned net, and the channels in each layer are divided into 15 groups (0.1x for each group). While for ResNet, we use standard ResNet50 (1.0x)  and ResNet18 (1.0x) as unpruned structures,
the channels in each layer are divided into 10 groups (0.1x for each group).

\noindent\textbf{DMCP training.}
As described in Section~\ref{sec:hpn_train}, the training pipeline of DMCP contains two phases: warmup and iterative training. 
The training is conducted on 16 Nvidia GTX 1080TI GPUs with a batch size of 1024.
Both MobileNetV2 and ResNet are trained for 40 epochs in total, the initial learning rate for both unpruned net and architecture parameters updating is 0.2 and reduced to 0.02 by cosine scheduler finally.

In the warmup phase, only the network weights are trained for 20 epochs using a variant of ``sandwich rule''.
In the iterative training phase, architecture parameters and unpruned net are both trained in a total of 20 epochs. The $\lambda_{reg}$ of budget regularization is set to 0.1 in all experiments. The tolerance ratio $\gamma$ is set to be 0.95 in all the experiments.
To make the explanation brief in the following sections, we use the shortened form of experiment settings. 
For example, MBV2 1.0x-59M means the unpruned net is MobileNetV2 1.0x with target FLOPs equals to 59M.

\noindent\textbf{Pruned network training.}
\label{sec:pruned_train}
The pruned networks are produced by Direct Sampling or Expected Sampling.
The details of the pruned model producing methods are illustrated in Section~\ref{sec:produce_model}.
Note that all pruned models are trained from scratch.
The training of pruned models is performed on 32 Nvidia GTX 1080TI GPUs with a batch size of 2048.
The pruned MobileNetV2 is trained for 250 epochs and pruned ResNet is trained for 100 epochs.
The initial learning rate for training all pruned models is first warming up from 0.2 to 0.8 within one epoch, then is reduced to 0 by cosine scheduler.

\subsection{Ablation Study}
\label{sec:ablation}

\begin{figure}[h]
    \centering
    \includegraphics[width=0.4\textwidth]{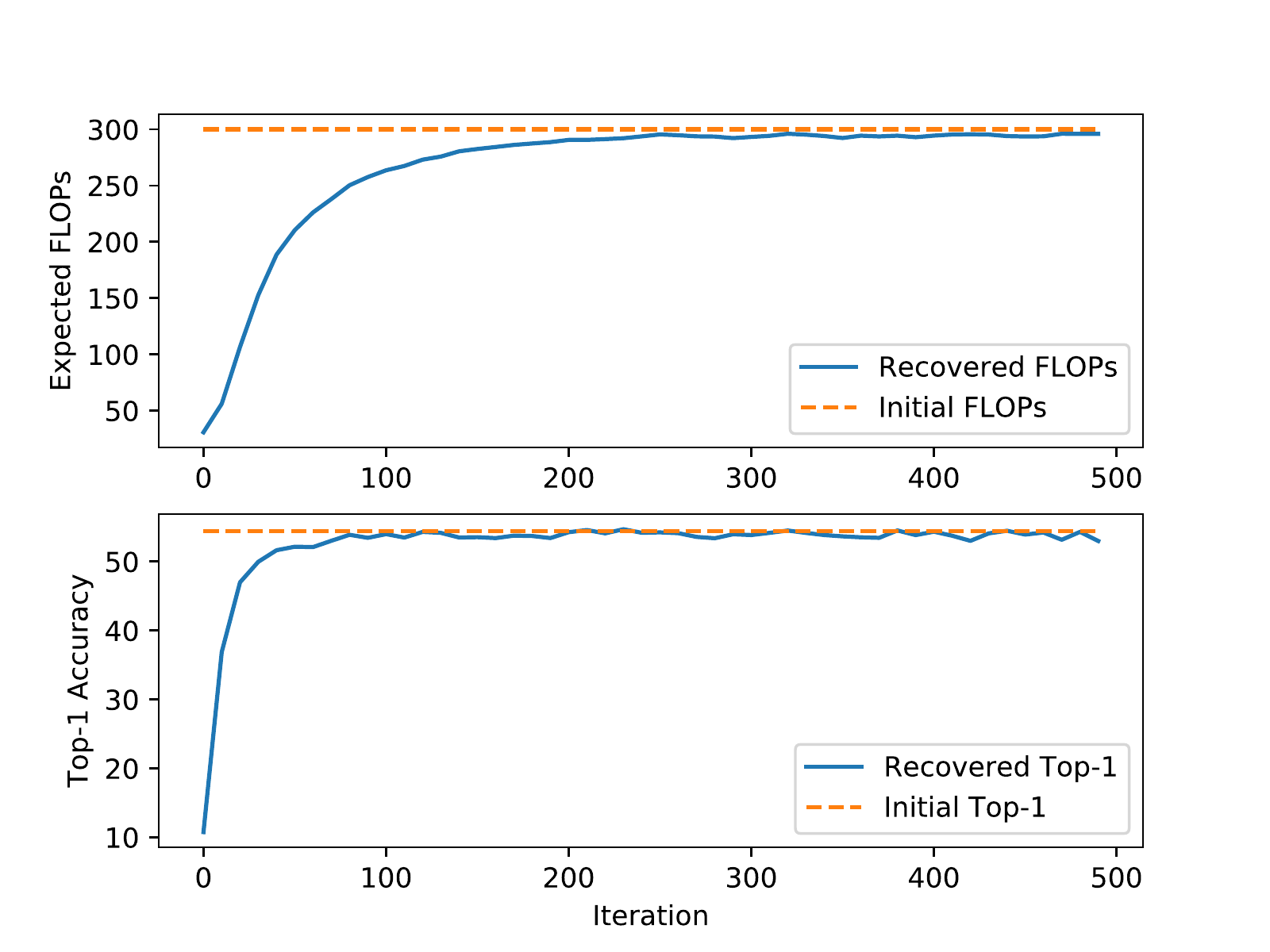}
    \caption{The recoverability of DMCP with pre-trained MobileNetV2 1.0x and randomly initialized architecture parameters.}
    \label{fig:recover}
\end{figure}

\noindent\textbf{Recoverability verification.}
One property of our method should have is that it should retain nearly all channels when searching on a pre-trained model without FLOPs constraint.
We use pre-trained MobileNetV2 1.0x and randomly initialize the architecture parameters.
Note that only the iterative training phase is performed. 
We freeze the weight of MobileNetV2 1.0x and trained architecture parameters with only task loss.
The result in Figure~\ref{fig:recover} shows that the FLOPs and top-1 training accuracy of our method can recover to those of the pre-trained model within 500 iterations.

\noindent\textbf{Expected sampling and Direct Sampling.}
As described in Section~\ref{sec:produce_model}, we can sample pruned models by Direct Sampling (DS) and Expected Sampling (ES).
We verify the effectiveness of two model producing methods on MobileNetV2-210M and ResNet50-1.1G.
We also train MobilenetV2 0.75x and ResNet50 0.5x, whose FLOPS is 210M and 1.1G respectively, as baselines for comparison. 
The performance of these two baselines are 70.4\% and 71.9\% separately.
For DS, we sample five models and the results are reported in Table~\ref{tb:direct_sample}.
The table shows that the performance of all models produced by DS is better than baseline models, which means the architecture parameters converge to a high-performance sub-space.
And the performance of model produced by ES is very close to the best model produced by DS, which shows the effectiveness of ES.
Besides, results from Table~\ref{tb:hypernet_scale} and Table~\ref{tb:warmup} also show the robustness of the ES.
For saving the cost of fine-tuning, we use the ES to produce a model if not indicated.

\begin{table}[h]
\centering
\begin{tabular}{c|c|cc}
\multirow{2}{*}{DMCP} &  \multirow{2}{*}{ES} & \multicolumn{2}{c}{DS} \\ 
\cline{3-4}
                         &    &  Highest  & Lowest  \\
\hline
MBV2 1.5x-210M & 72.2 & 72.4 & 70.6  \\ 
\hline
Res50 1.0x-1.1G & 74.1 & 74.0 & 72.3 \\ 
\end{tabular}
\caption{Performance of pruned model produced by Direct Sampling (DS) and Expected Sampling (ES). 
``Highest'' and ``Lowest'' means the best and worst performance among 5 models sampled by Direct Sampling. MBV2 is short for MobileNetV2 and Res50 is short for ResNet 50.
}
\label{tb:direct_sample}
\end{table}

\noindent\textbf{The scale of the unpruned network.}
In this section, we evaluated the influence of scaling the unpruned network.
We use two scales of MobileNetV2, i.e. MobileNetV2 1.0x and MobileNetV2 1.5x, as unpruned network, and prune them into 59M and 210M FLOPs.
Note that in the experiments, the channels in each layer are divided into 10 groups to maintain the same group size.
The results showed in Table~\ref{tb:hypernet_scale} indicate that our method is not sensitive to the unpruned network scale.
Using the larger unpruned network can lead to a little bit better performance.
So we use MobileNetV2 1.5x and ResNet50 1.0x as our default unpruned network in the remaining paper.

\begin{figure}[ht]
    \centering
    \includegraphics[width=0.45\textwidth]{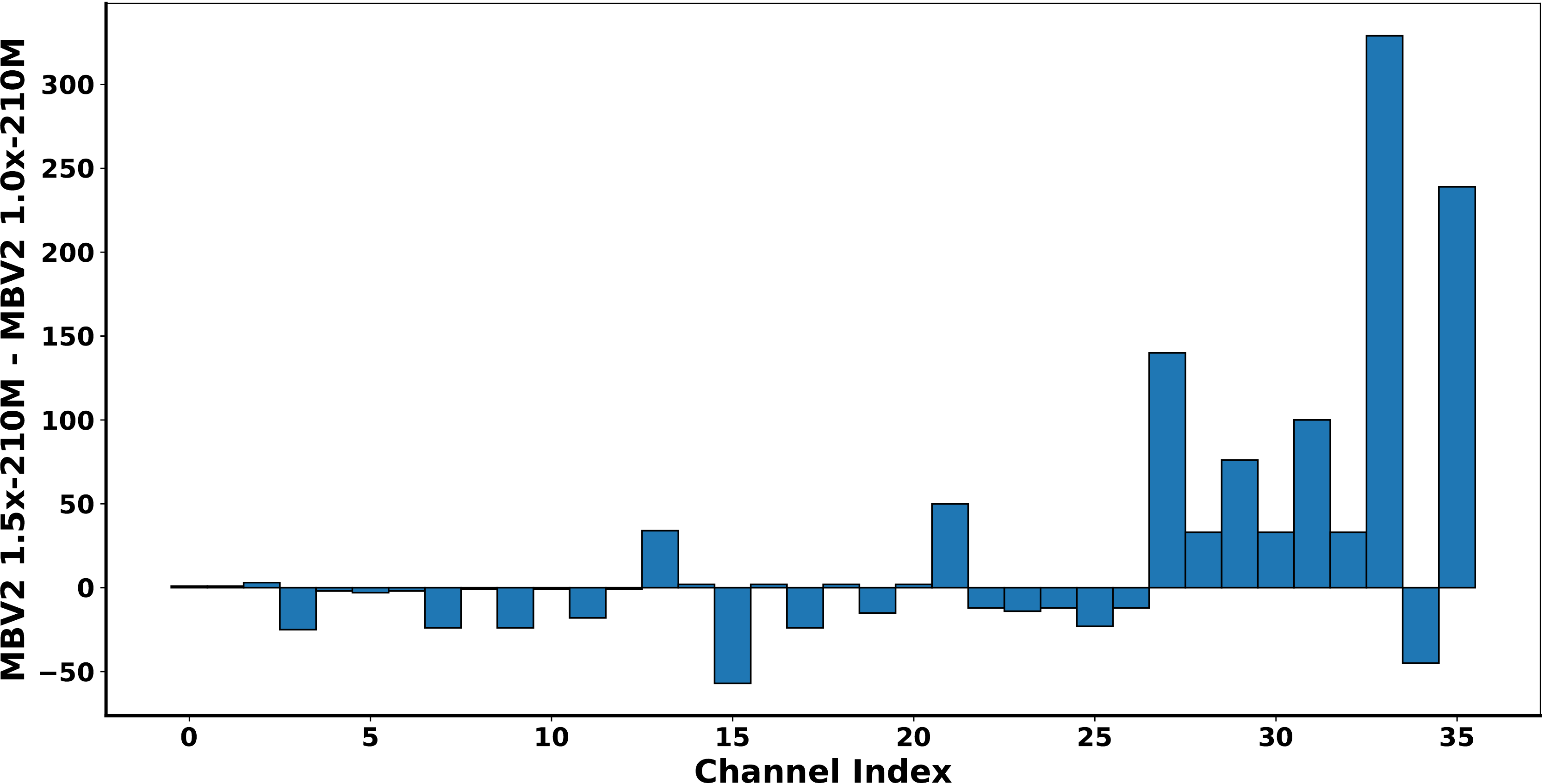}
    \caption{The difference between two pruned models from MBV2 1.5x-210M and MBV2 1.0x-210M. The x-axis indicates the layer index and the y-axis is the difference computed by subtracting the number of channels each layer in MBV2 1.0x-210M from that in MBV2 1.5x-210M.
    \vspace{-1em}}
    \label{fig:compare}
\end{figure}

We also visualize the difference computed by subtracting the number of channels each layer in MBV2 1.0x-210M from that in MBV2 1.5x-210M in Figure~\ref{fig:compare}.
From the figure, we can observe that MBV2 1.0x-210M tends to retain more channels in shallow layers while MBV2 1.5x-210M retains more channels in deep layers, even they only have a tiny difference in accuracy.
This indicates that there exist multiple local minima in the search space of channel pruning.

\begin{table}[h]
\centering
\begin{tabular}{c|c|cc}
\multirow{2}{*}{DMCP} & \multirow{2}{*}{ES} & \multicolumn{2}{c}{DS} \\ 
\cline{3-4}
                           &                           &  Highest  & Lowest  \\
\hline
MBV2 1.5x-59M & 62.7 & 62.9 & 60.8 \\ 
MBV2 1.0x-59M & 62.6 & 62.6 & 60.6 \\ 
\hline 
MBV2 1.5x-210M  & 72.2 & 72.4 & 71.4\\ 
MBV2 1.0x-210M  & 71.8 & 72.0 & 70.4 \\ 
\end{tabular}
\caption{The performance of pruned models in 59M and 210M FLOPs level on MobileNetV2 (MBV2) with different unpruned network scale. 
}
\label{tb:hypernet_scale}
\end{table}

\noindent\textbf{Influence of warmup phase.}
We train MobileNetV2-210M with and without the warmup phase and evaluate their performance of the corresponding pruned models. 
To keep other settings the same, we double the epochs of the iterative training phase for the experiment without warming up.
In the setting without warming up, the models are trained for 100 epochs, the initial learning rate and the scheduler in the first 50 epochs are the same as the warmup phase.
The results in Table~\ref{tb:warmup} shows that using warmup leads to better performance.
One possible reason is that using warmup makes the weights trained more sufficiently before updating architecture parameters, which makes weights more discriminable and prevents architecture parameters from trapping into bad local minima.

\begin{table}[h]
\centering
\begin{tabular}{c|c|c|cc}
\multirow{2}{*}{DMCP}                 &\multirow{2}{*}{Warmup} & \multirow{2}{*}{ES} &\multicolumn{2}{c}{DS} \\ 
\cline{4-5}
                                       &                       &                          &  Highest  & Lowest         \\
\hline
\multirow{2}{*}{MBV2-210M} 
                                       & \checkmark             & 72.2                     & 72.4         & 71.4              \\ 
                                       & \ding{55}              & 71.4                     & 71.2         & 70.5              \\ 
\end{tabular}
\caption{The influence of using warmup or not. 
MBV2 is short for MobileNetV2. 
}
\label{tb:warmup}
\end{table}

\noindent\textbf{Impact of the variant sandwich rule.} 
We ablate the impact of the sandwich rule in the MobileNetV2-210M setting.
The original sandwich rule and our variant sandwich rule are adopted solely in DMCP for comparison.
The results are tabulated in Table~\ref{tb:scheduled_random}.
We can see that using the variant sandwich rule leads to better performance.
The possible reason is that the weights corresponding to higher probability will be optimized better by the variant sandwich rule.
And in these weights, each of them will be optimized better to represent their true importance with less influence of other weights.
Thus, when updating architecture parameters, the ``competition'' is mainly centered on them, which makes updating more accurate.

\begin{table}[h]
\centering
\begin{tabular}{c|c|c}
DMCP   & Sandwich rule & Top-1 \\
\hline
\multirow{2}{*}{MBV2-210M} 
                                       & original & 71.5  \\ 
                                       & our variant & 72.2  \\ 
\end{tabular}
\caption{The influence of using the variant sandwich rule or not. 
}
\label{tb:scheduled_random}
\end{table}

\noindent\textbf{Training scheme.}
We verify the effectiveness of the updating scheme.
We conduct three experiments on MobileNetV2-59M.
All experiments use the same setting in the warmup phase, while the settings in the iterative training phase are as follows: 
In the first experiment, we only update architecture parameters with respect to budget regularization (FLOPs loss); 
in the second experiment, we only update architecture parameters with respect to both budget regularization and task loss;
and in the last experiment, we update both unpruned net and architecture parameters with respect to the full loss function.
The results are shown in Table~\ref{tb:update_scheme}.
The first experiment is a naive baseline of FLOPs guided pruning.
The layers with the same FLOPs may be pruned to the same extent.
The result is far worse than the other experiments.
Comparing with the first experiment and the second experiment, we know that the task loss can help to discriminate the importance of different layer even they have same FLOPs.
Finally, by comparing with the last two experiments, we can conclude that when architecture parameters changed, the weights should also be adapted.

\begin{table}[]
\centering
\begin{tabular}{cc|c|c}
\multicolumn{2}{c|}{Arch. Params. updating}& \multirow{2}{*}{Weight updating} &  \multirow{2}{*}{Top-1} \\ 
\cline{1-2}
 FLOPs loss    & Task loss       &                                &                                 \\
\hline
\checkmark    &                 &                                &   52.1                          \\
\checkmark    & \checkmark      &                                &   61.5                      \\
\checkmark    & \checkmark      & \checkmark                     &   62.7                          \\
\end{tabular}
\caption{The influence of different components in iterative training phase.
The experiments are conducted with MobileNetV2 and all the target FLOPs are 59M.
The cell without check-mark means the corresponding component is not used during training.
\vspace{-1.5em}
}
\label{tb:update_scheme}
\end{table}

\begin{table}[]
    \centering
    \begin{tabular}{c|lccc}
    Group  & Model & FLOPs & Top-1 & $\Delta$ Top-1\\
    \hline
    \multirow{20}{*}{MBV2} 
        
                            & Uniform 1.0x & 300M & 72.3 & -\\
                            & Uniform 0.75x & 210M & 70.1 & -2.2\\
                            & Uniform 0.5x & 97M & 64.8 & -7.5\\
                            & Uniform 0.35x & 59M & 60.1 & -12.2\\
                            \cline{2-5}             
                            & \multirow{3}{*}{MetaPruning\cite{liu2019metapruning}} & 217M & 71.2 & -0.8\\
                            &                                          & 87M & 63.8 & -8.2 \\
                            &                                          & 43M & 58.3 & -13.7 \\
                            \cline{2-5}             
                            & AMC\cite{he2018amc} & 211M & 70.8 & -1.0\\
                            \cline{2-5} 
                            & \multirow{2}{*}{AutoSlim\footnotemark\cite{yu2019autoslim}
                            \textbf{*}} 
                            & 300M & 74.2 & +2.4\\
                            &                                       & 211M & 73.0 & +1.2\\
                            \cline{2-5}             
                            & \multirow{6}{*}{DMCP} & 300M & 73.5 & +1.2\\
                            &                                  & 211M & 72.2 & -0.1 \\
                            &                                  & 97M & 67.0 & -5.3 \\
                            &                                  & 87M & 66.1 & -6.2 \\
                            &                                  & 59M & 62.7 & -9.6 \\
                            &                                  & 43M & 59.1 & -13.2 \\  
                            \cline{2-5}
                            & \multirow{2}{*}{DMCP\textbf{*}}    & 300M & 74.6 & +2.3\\
                            &                                           & 211M & 73.5 & +1.2\\   
    \hline 
    \multirow{3}{*}{Res18}  
                            & Uniform 1.0x & 1.8G & 70.1 & -\\ 
                            \cline{2-5}
                            
                            & FPGM\cite{DBLP:journals/corr/abs-1811-00250} & 1.04G & 68.4 & -1.9\\
                            \cline{2-5}
                            & DMCP
                             & 1.04G & 69.2 & -0.9 \\
                            
    \hline

    \multirow{17}{*}{Res50}    
                                        & Uniform 1.0x & 4.1G & 76.6 & -\\ 
                                        & Uniform 0.85x & 3.0G & 75.3 & -1.3\\ 
                                        & Uniform 0.75x & 2.3G & 74.6 & -2.0\\ 
                                        & Uniform 0.5x & 1.1G & 71.9 & -4.7\\ 
                                        & Uniform 0.25x & 278M & 63.5 & -13.1 \\ 
    \cline{2-5}
                                        & FPGM\cite{DBLP:journals/corr/abs-1811-00250} & 2.4G & 75.6 & -0.6 \\  
    \cline{2-5}
                                        & SFP~\cite{he2018progressive} & 2.4G & 74.6 & -2.0 \\  
    \cline{2-5}
                                        & \multirow{3}{*}{MetaPruning\cite{liu2019metapruning}} & 3.0G & 76.2 & -0.4\\
                                        &                                        & 2.3G & 75.4 & -1.2 \\
                                        &                                        & 1.1G & 73.4 & -3.2 \\
    \cline{2-5}
                                        & \multirow{3}{*}{AutoSlim\cite{yu2019autoslim}*} & 3.0G & 76.0 & -0.6\\
                                        &                                        & 2.0G & 75.6 & -1.0\\
                                        &                                        & 1.1G & 74.0 & -2.6\\
    \cline{2-5}
                                        & \multirow{4}{*}{DMCP} & 2.8G & 76.7 & +0.1 \\
                                        &                                & 2.2G & 76.2 & -0.4\\
                                        &                                & 1.1G & 74.4 & -2.2\\
                                        &                                & 278M & 66.4 & -10.0\\ 
    
    \end{tabular}
    
\caption{Performance of different models on ImageNet dataset with different FLOPs settings. $\Delta$ Top-1 column list the accuracy improvement compared with unpruned baseline model (1.0$\times$) reported in their original work, and our baseline is indicated by ``-''. ``$\alpha\times$'' means each layer in baseline model is scaled by $\alpha$. The groups marked by \textbf{*} indicate the pruned model is trained by slimmable method proposed in~\cite{yu2019autoslim} \vspace{-2em}}
\label{tb:models}
\end{table}

\subsection{Comparison with state-of-the-art}
\label{sec:compare}
In this section, we compare our method with various pruning methods, including reinforcement learning method AMC~\cite{he2018amc}, evolution method MetaPruning~\cite{liu2019metapruning}, one-shot method AutoSlim~\cite{yu2019autoslim}, and traditional channel pruning methods SFP~\cite{he2018progressive} and FPGM~\cite{DBLP:journals/corr/abs-1811-00250}.
The training settings of our method in all FLOPs settings are illustrated in Section~\ref{sec:imp_detail}, and our pruned models are sampled by Expected Sampling.

All methods are evaluated on MobileNetV2, ResNet18, and ResNet50, in each type of model, we trained a set of baseline  model with setting~\ref{sec:pruned_train} for comparison.

From the Table~\ref{tb:models}, we can see that our method outperforms all other methods under the same settings, which show the superiority of our method.
Note that AMC, MetaPruning and our method train the pruned model from scratch by standard hard label loss. 
While AutoSlim adopts a in-place distillation method in which the pruned network share weights with unpruned net and mimic the output of the unpruned net.
To fairly compare with AutoSlim, we also train our pruned model with the slimmable training method.
Results show that this training method can further boost the performance, and our method surpasses AutoSlim in different FLOPs models.

\footnotetext{Training settings of baseline and pruned models are different.}
\section{Conclusion}
In this paper, we propose a novel differentiable method for channel pruning, named Differentiable Markov Channel Pruning (DMCP), to solve the defect of existing methods that they need to train and evaluate a large number of sub-structures.
The proposed method is differentiable by modeling the channel pruning as the Markov process, thus can be optimized with respect to task loss by gradient descent.
After optimization, the required model can be sampled from the optimized transitions by a simple Expected Sampling and trained from scratch.
Our method achieves state-of-the-art performance with ResNet and MobileNet V2 on ImageNet in various FLOPs settings.

{\small
\bibliographystyle{ieee_fullname}
\bibliography{dmcp}
}

\newpage
\appendix
\section{Visualization}
\label{sec:vis}
\subsection{FLOPs Distribution of the Pruned Model}
\begin{figure}[h]
    \centering
    \includegraphics[width=0.4\textwidth]{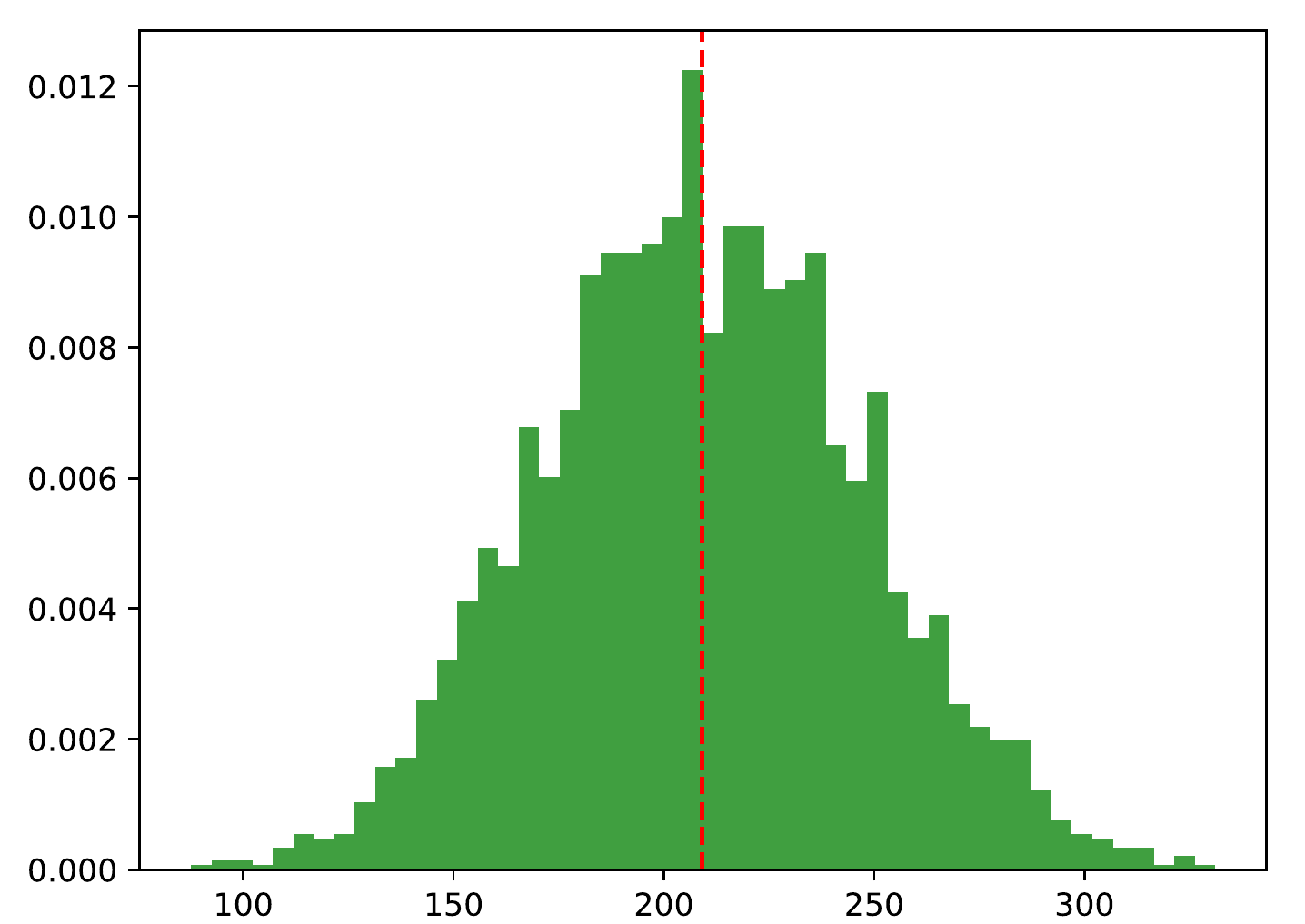}
    \caption{The FLOPs distribution of 3000 structures sampled from MobileNetV2-210M by Markov process. The x-axis is the MFLOPs and the y-axis is the frequency. The red dashed line is the mean of the FLOPs of 3000 sampled structures. The FLOPs of the unpruned network is 672M.}
    \label{fig:flops_stats}
\end{figure}

We sample 3000 structures from the trained MobileNetV2-210M via the Markov process, and the distribution of their FLOPs is showed in Figure~\ref{fig:flops_stats}. 
From the figure, we can find that the mean of FLOPs lies around 210M, which means that the expected FLOPs converged to the desired budget 210M.

\begin{figure*}[h]
\centering
\begin{tabular}{ccc}
\includegraphics[width=0.32\linewidth]{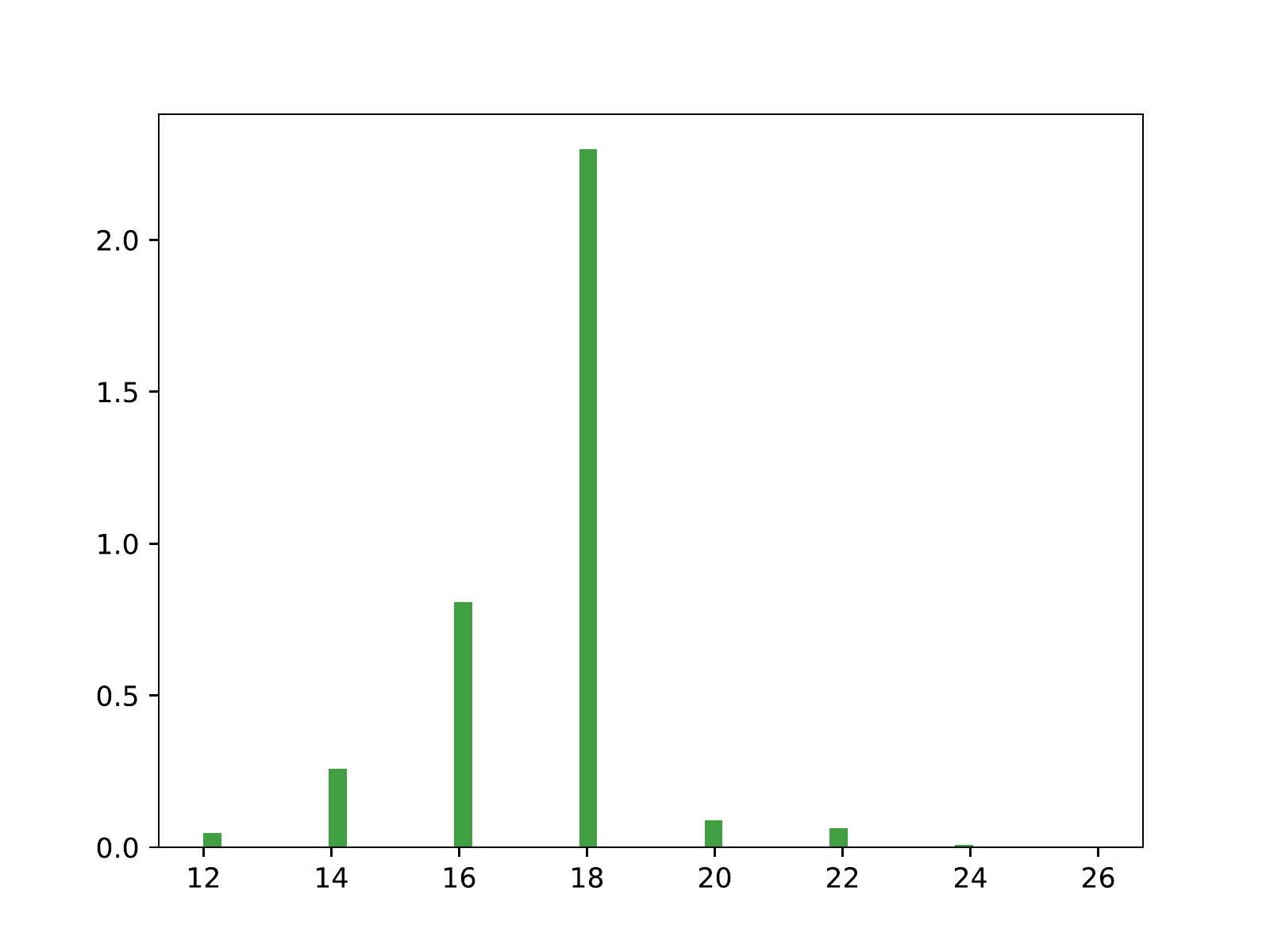} &
\includegraphics[width=0.32\linewidth]{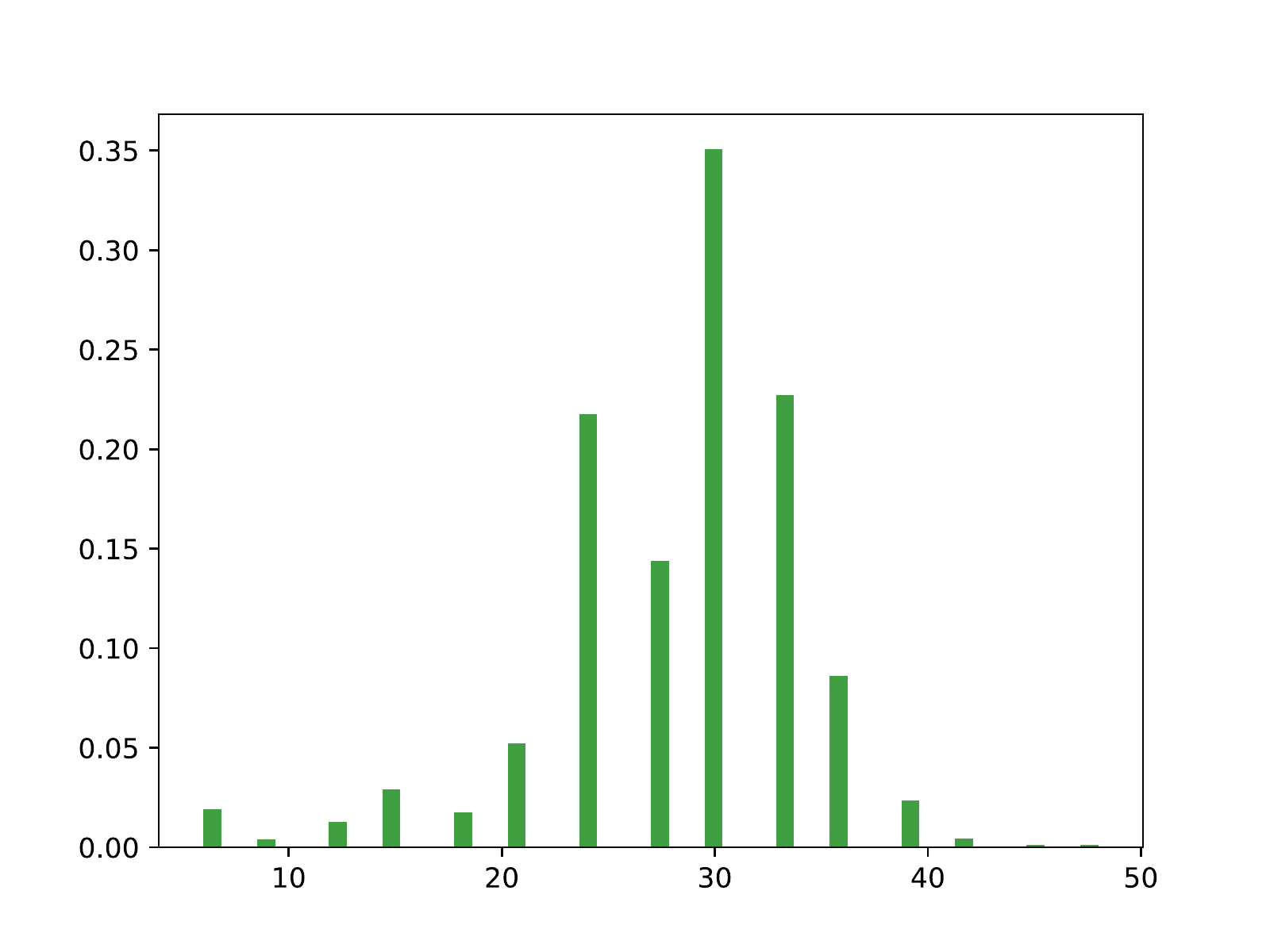} & 
\includegraphics[width=0.32\linewidth]{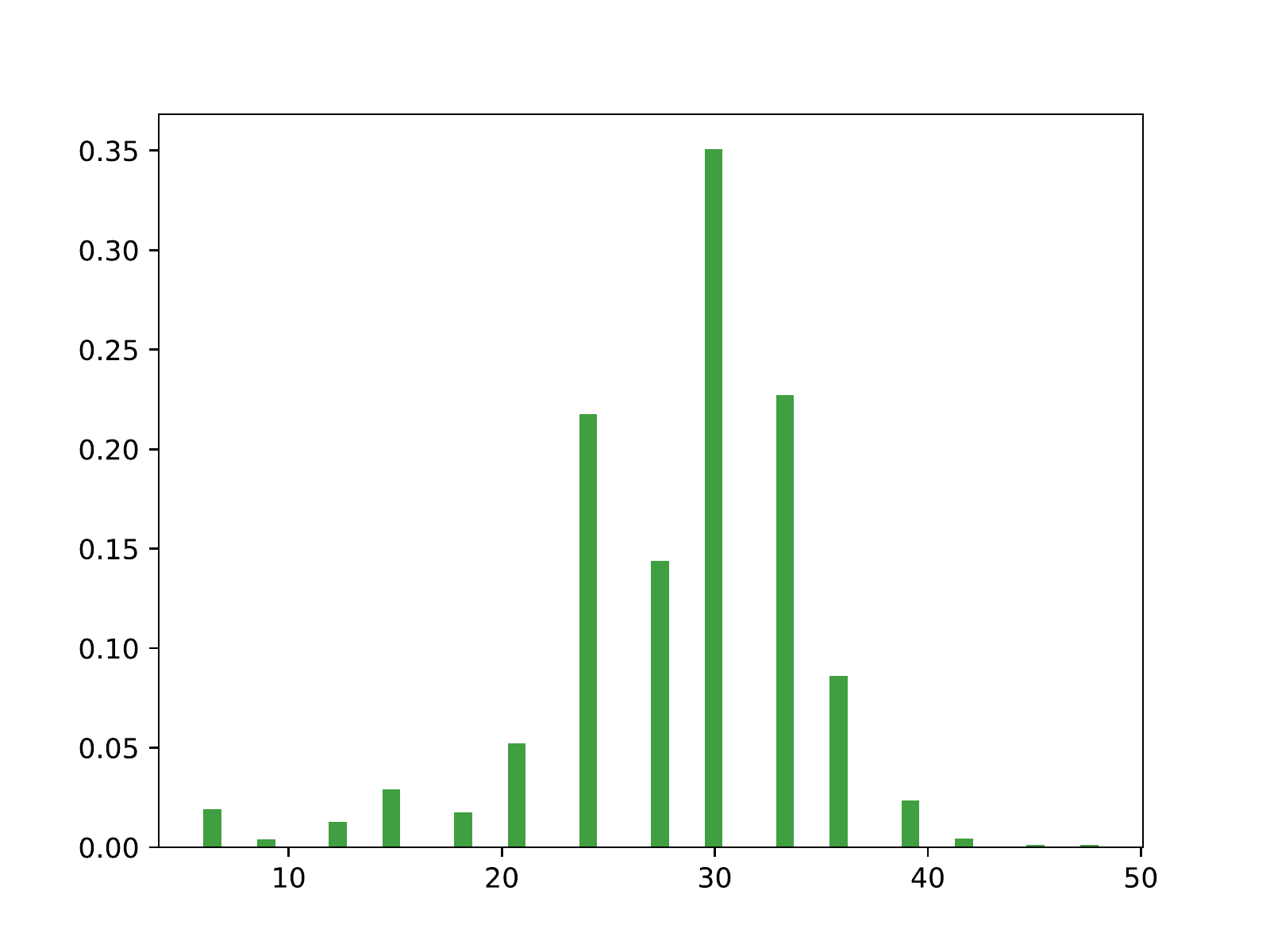} \\
\hspace{0.2cm}\scriptsize  LinearBottleneck2\_2.conv3 & \hspace{0.2cm}\scriptsize LinearBottleneck3\_1.conv3 &
\hspace{0.2cm}\scriptsize LinearBottleneck3\_3.conv3 
\\
\includegraphics[width=0.32\linewidth]{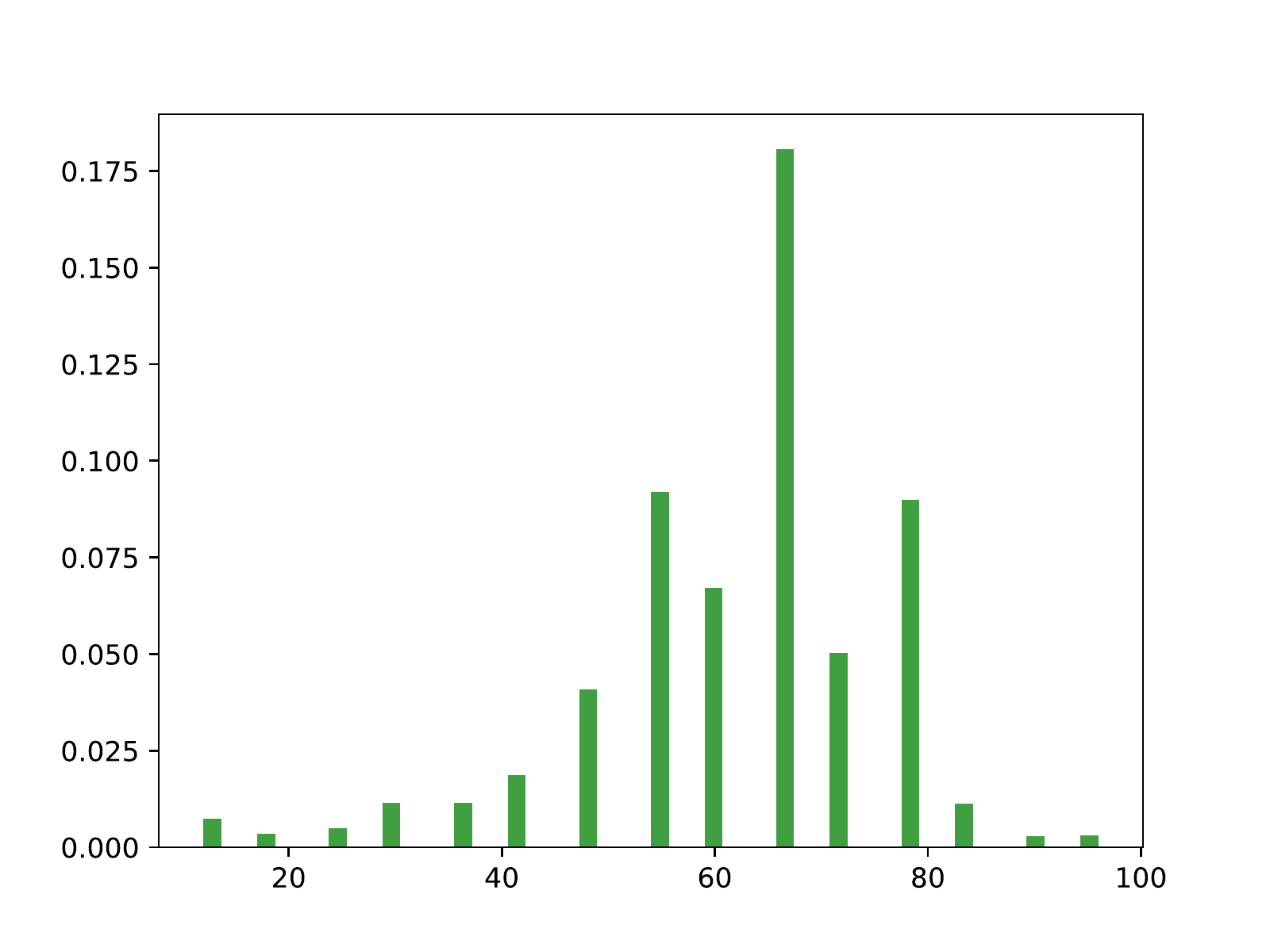} & \includegraphics[width=0.32\linewidth]{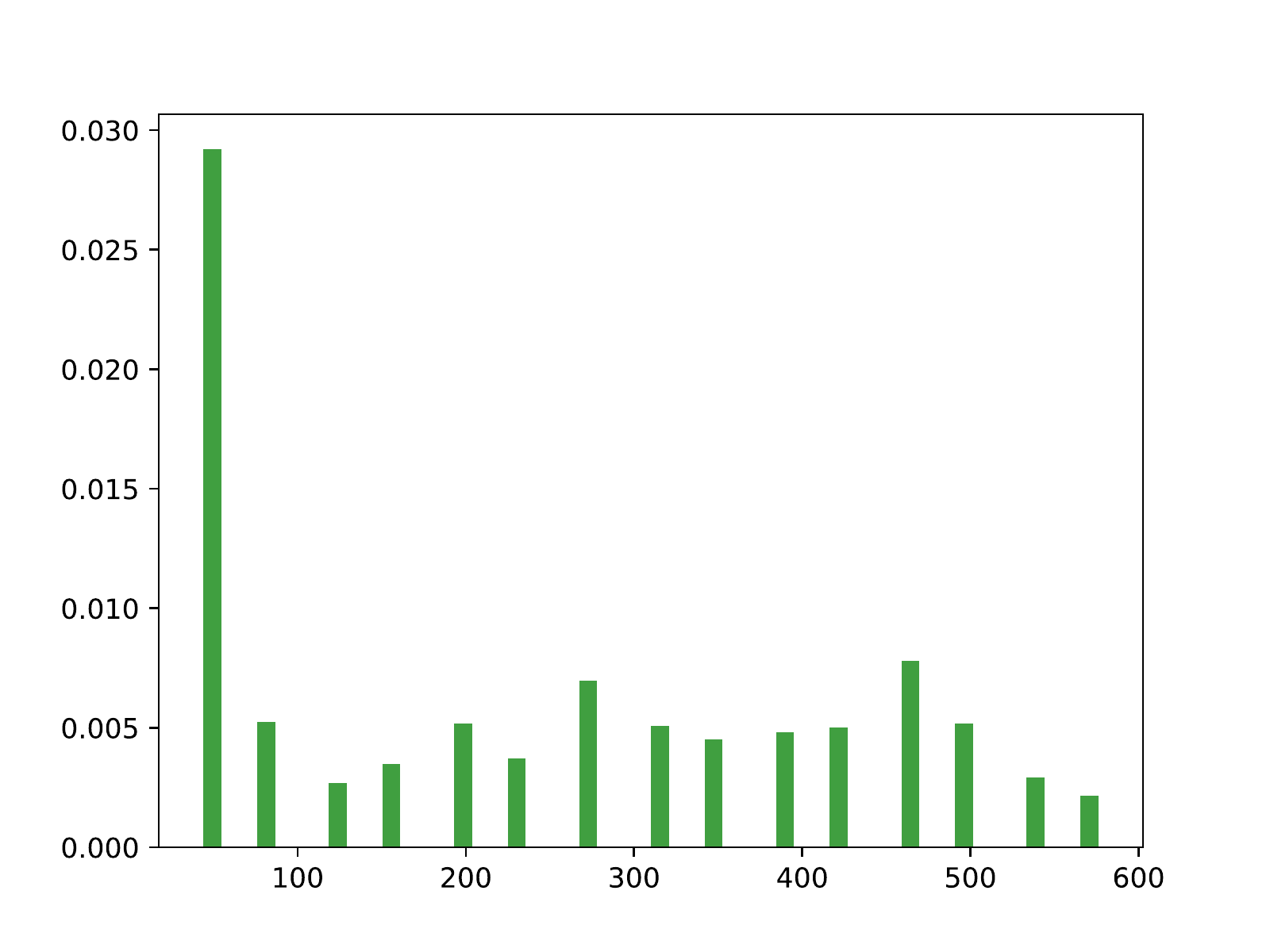} & 
\includegraphics[width=0.32\linewidth]{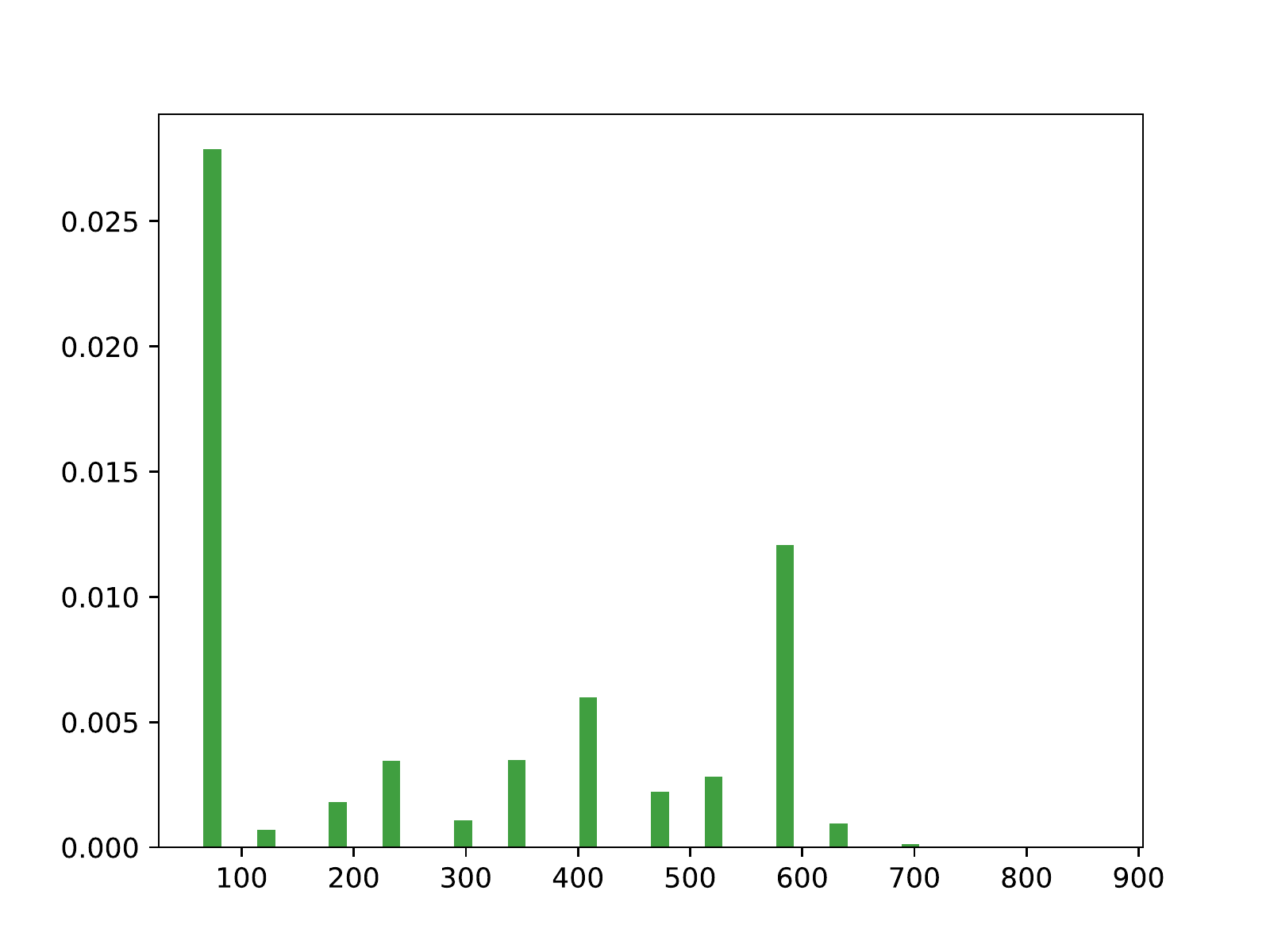}
\\
\hspace{0.2cm}\scriptsize  LinearBottleneck4\_0.conv3 & \hspace{0.2cm}\scriptsize  LinearBottleneck4\_4.conv1&
\hspace{0.2cm}\scriptsize  LinearBottleneck5\_2.conv1
\\
\includegraphics[width=0.32\linewidth]{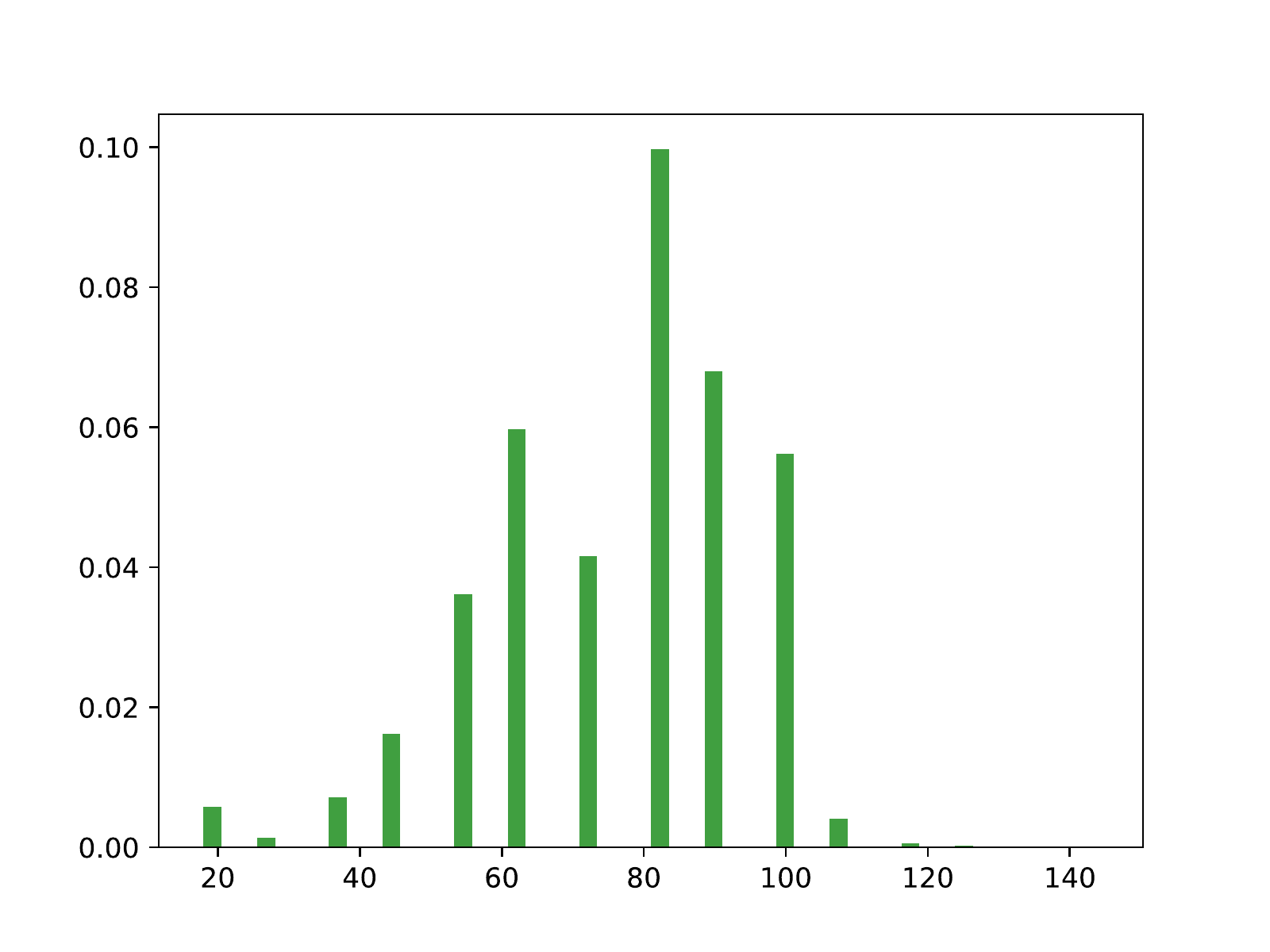} & \includegraphics[width=0.32\linewidth]{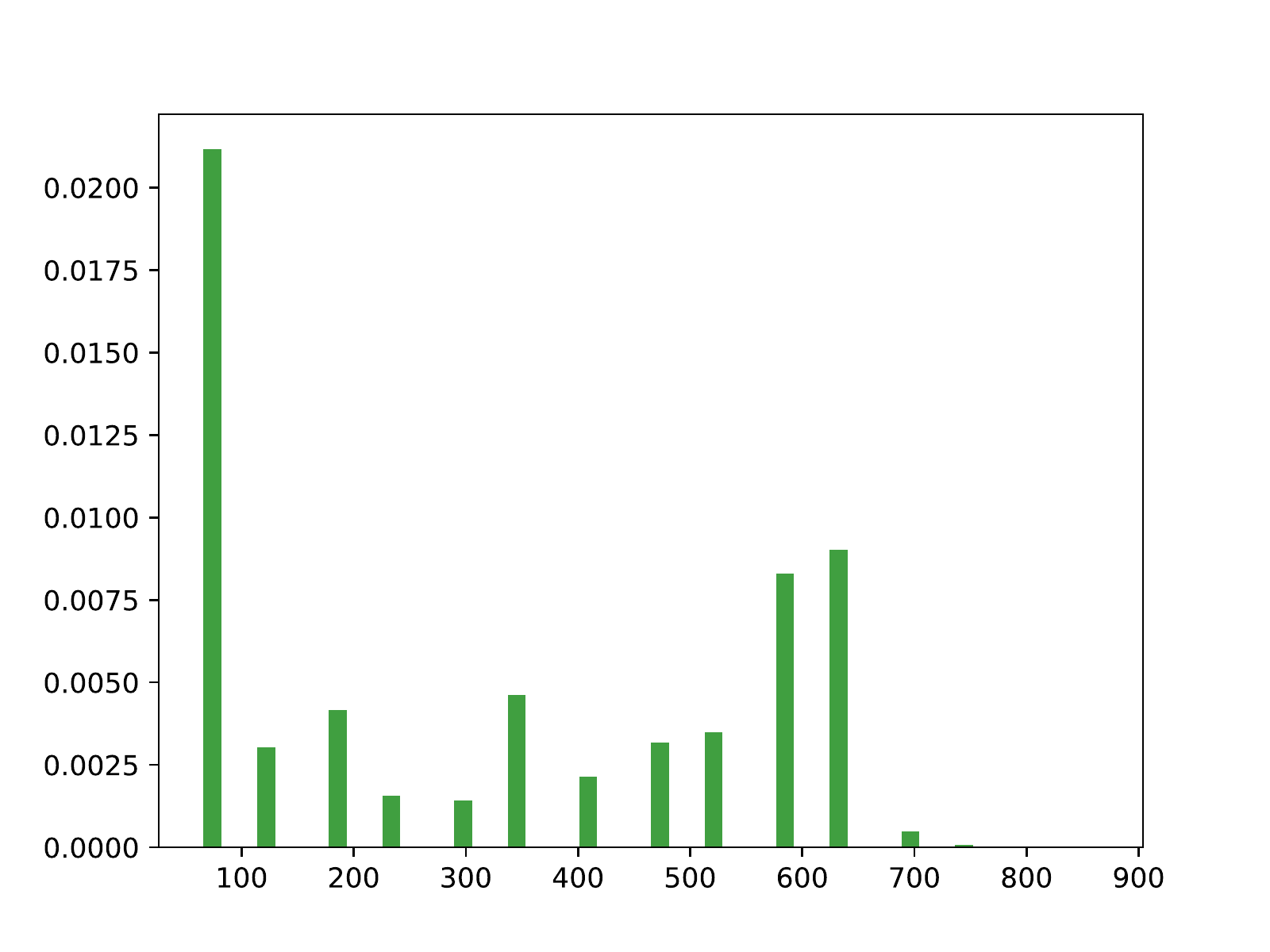} & 
\includegraphics[width=0.32\linewidth]{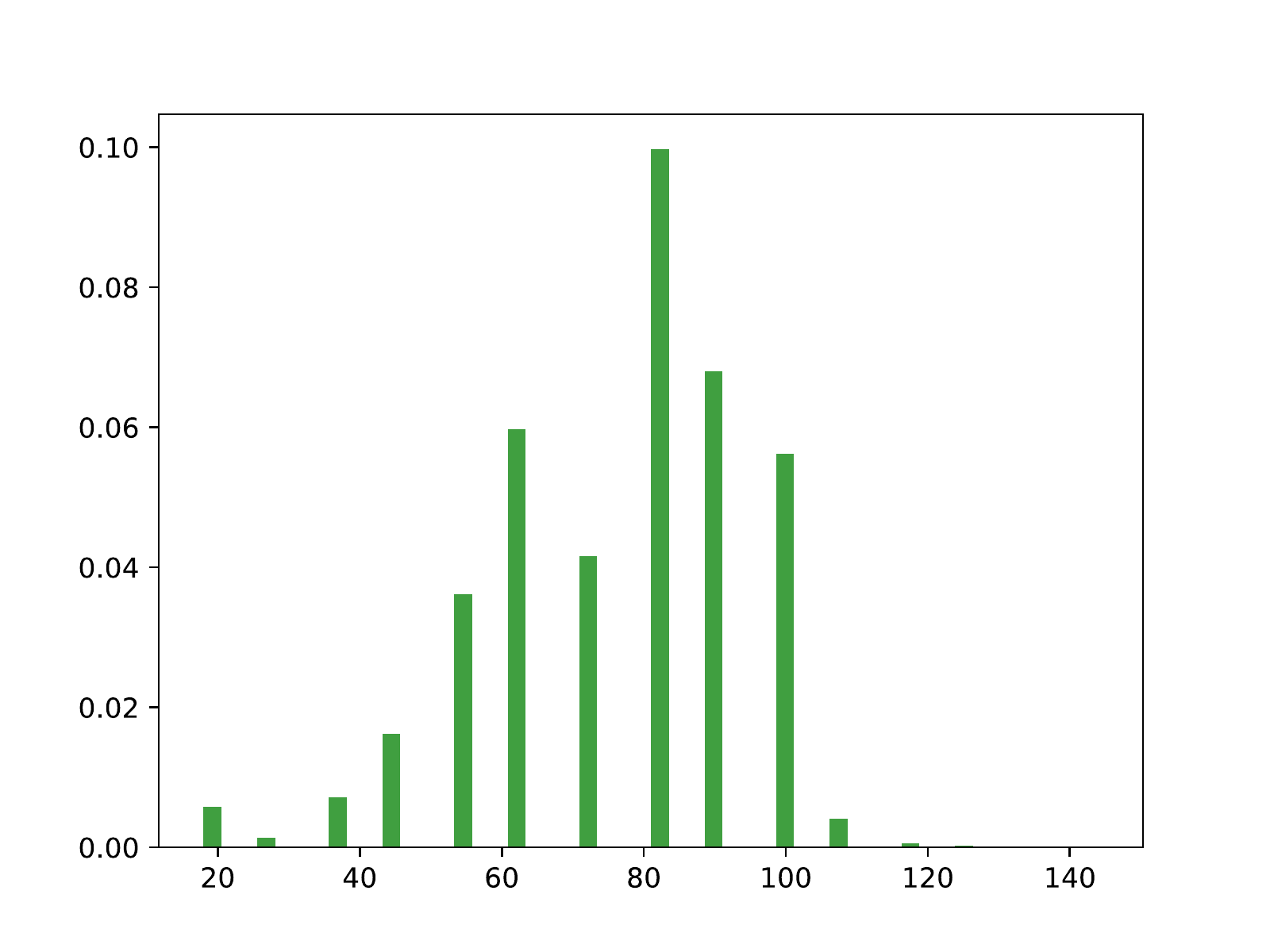}
\\
\hspace{0.2cm}\scriptsize  LinearBottleneck5\_2.conv3 & \hspace{0.2cm}\scriptsize  LinearBottleneck5\_3.conv1&
\hspace{0.2cm}\scriptsize  LinearBottleneck5\_3.conv3
\\
\includegraphics[width=0.32\linewidth]{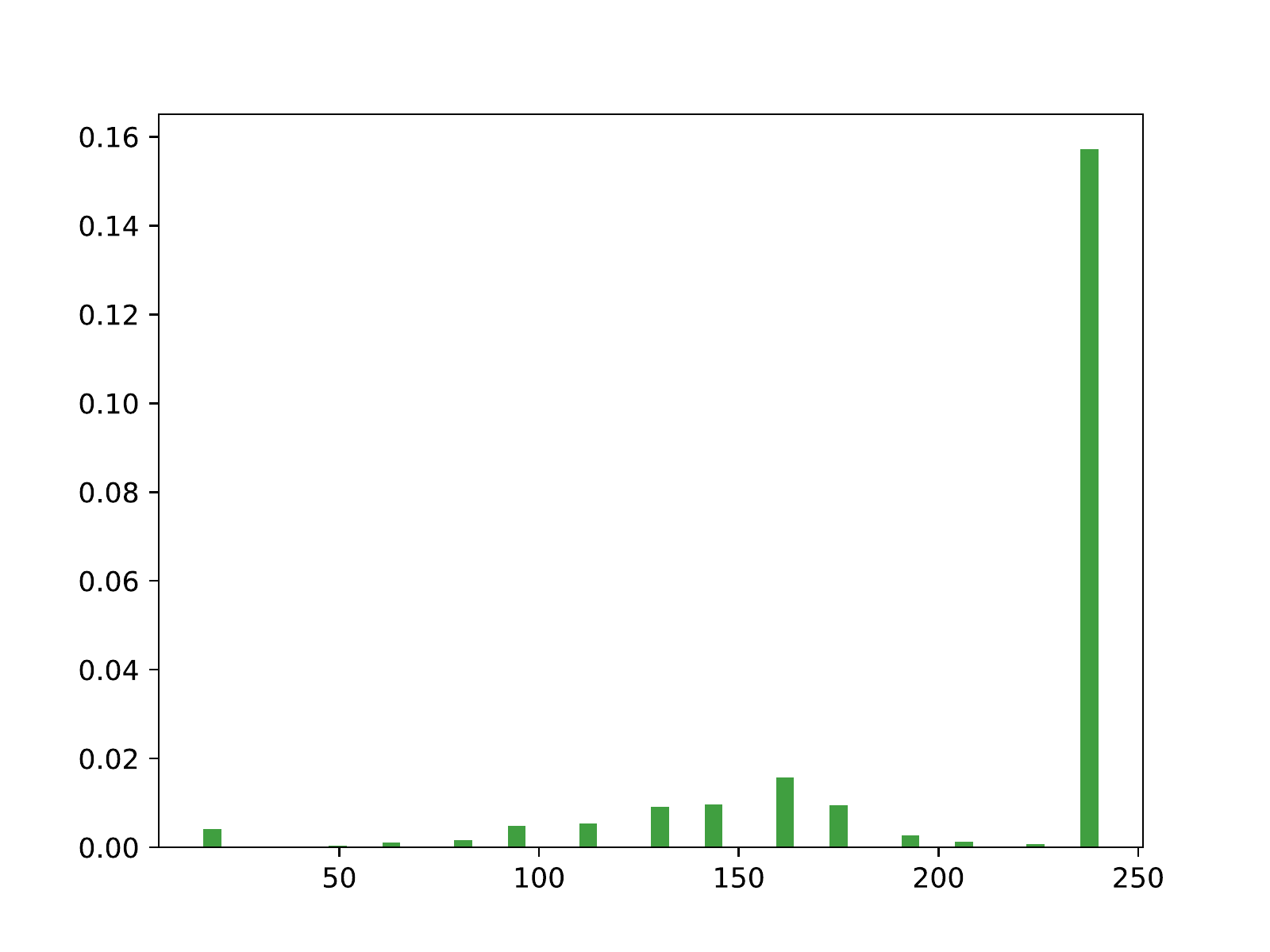} & \includegraphics[width=0.32\linewidth]{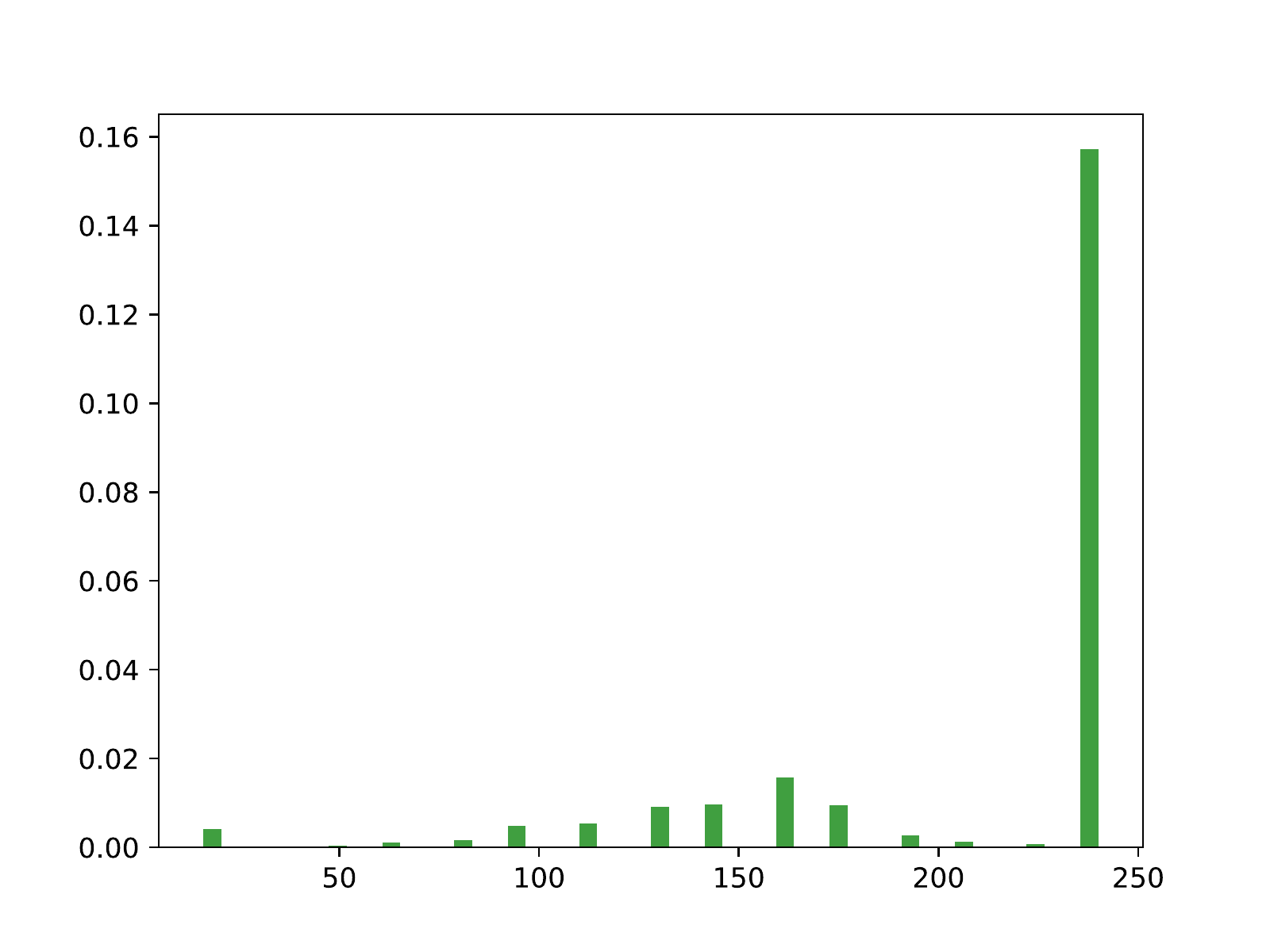} & 
\includegraphics[width=0.32\linewidth]{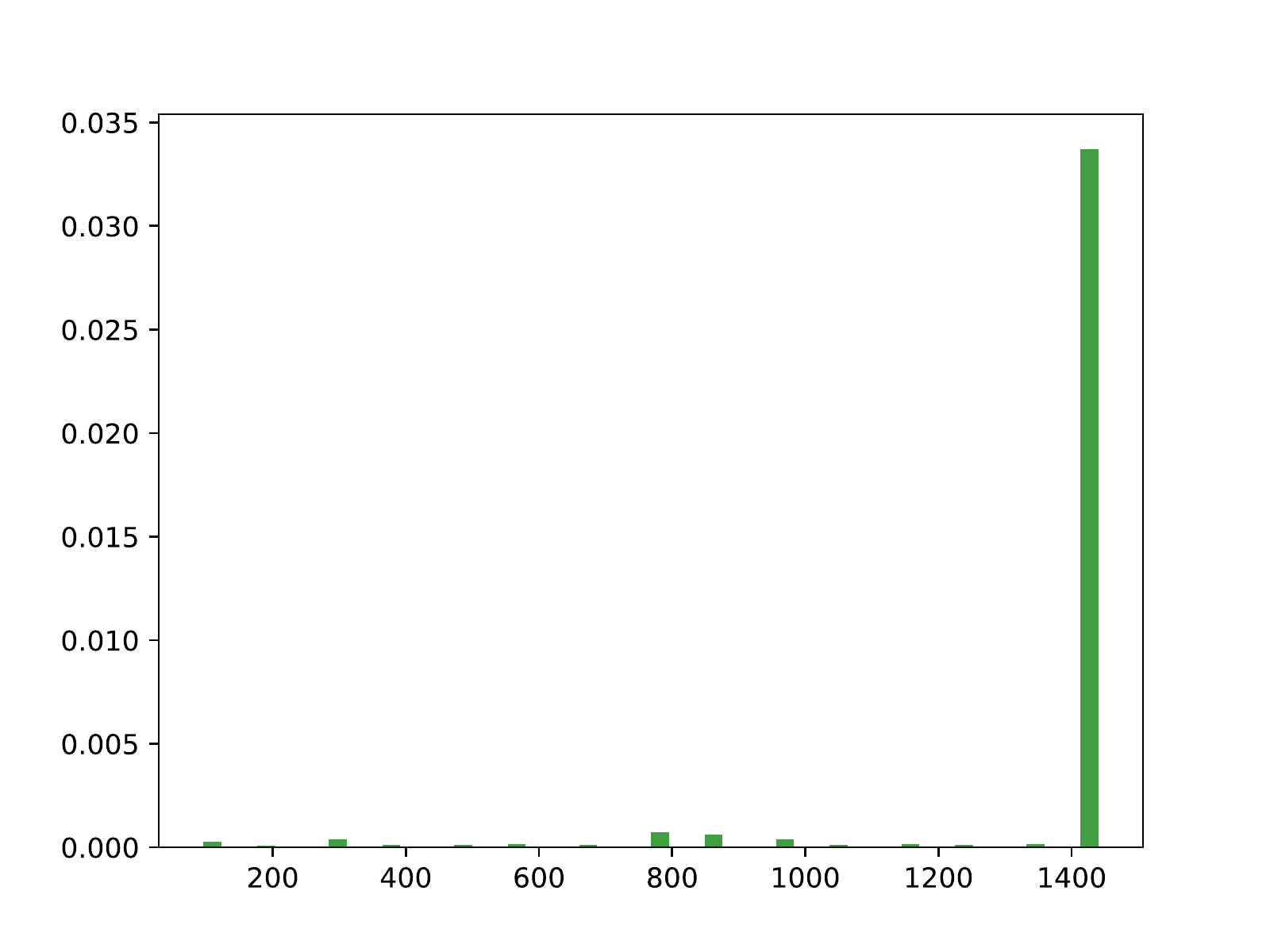}
\\
\hspace{0.2cm}\scriptsize  LinearBottleneck6\_2.conv3 & \hspace{0.2cm}\scriptsize  LinearBottleneck6\_3.conv1&
\hspace{0.2cm}\scriptsize  LinearBottleneck7\_1.conv1
\\
\end{tabular}
\caption{The channel distribution of 12 layers sampled from different blocks in MobileNetV2-210M. The y-axis is the frequency and the x-axis is the number of channels. Note that we divided channels each layer into 15 groups.}
\label{fig:chanel_dist}
\end{figure*}

\subsection{The Channel Distribution of Pruned Layers. }
In this section, we examine the channel distribution in each layer of the pruned model. 
We sample 3000 models from MobileNetV2-210M whose FLOPs are within the desired budget (210M) by Markov process. 
Figure~\ref{fig:chanel_dist} shows the channel distribution of 12 layers sampled from different blocks.
From the figure, we can observe that the number of channel in most layers follows an uni-modal distribution, 
and some layers choose to retain all the channels (e.g. LinearBottleneck6 and 7).

\section{Comparison between using warm-up and using pre-trained  model}
As described in Section 3.2, the warm-up is performed by only running stage 1 that updating the weights of the unpruned network by our proposed variant sandwich rule, which makes the channel group more  “important”  than the  channel group right after it, providing a good initialization for iterative training. However, using a pre-trained model cannot provide initialization with the property. 
Our experiment also shows the superiority of using warm-up, on DMCP-MBV2 with 210M FLOPs by replacing warm-up with using pre-trained models, a 0.6\% accuracy drop was observed.

\section{Modeling architecture parameters as independent Bernoulli variables}
Given a layer with $C$ channels, the solution space of our method is $O(C)$, by modeling architecture parameters as Bernoulli variables, the solution space becomes $O(2^{C})$, as there are $2^{C}$ possible channel combinations, which makes it much harder to optimize.
To demonstrate our analysis, we experiment on MobileNet-v2 with 210M FLOPs, by replacing Markov modeling with Bernoulli Modelling of architecture parameters, the performance of the pruned model is 70.1\%, which is 2.3\% lower than DMCP.

\end{document}